\pdfoutput=1

\documentclass[11pt]{article}

\usepackage[preprint]{acl}

\usepackage{times}
\usepackage{latexsym} 
\usepackage{booktabs}
\usepackage{subfiles}
\usepackage{graphicx}
\usepackage{array}
\usepackage{amsmath}   
\usepackage{amssymb}
\usepackage{amsmath}
\usepackage[T1]{fontenc}

\usepackage[utf8]{inputenc}

\usepackage{microtype}
\usepackage{subfiles}
\usepackage{inconsolata}

\usepackage{graphicx}
\usepackage{pifont}
\usepackage{xcolor}

\usepackage[most]{tcolorbox} 
\usepackage{xcolor}

\newtcolorbox{promptbox}[1]{%
  enhanced,
  colback=blue!5,
  colframe=blue!60!black,
  coltitle=white,
  fonttitle=\bfseries,
  title=#1, 
  attach boxed title to top left={xshift=6pt,yshift=-2pt},
  boxed title style={
    colback=blue!60!black,
    colframe=blue!60!black,
    arc=4pt,
    left=6pt,right=6pt,top=3pt,bottom=3pt
  },
  arc=6pt,
  boxrule=0.8pt,
  left=10pt,right=10pt,top=8pt,bottom=8pt,
  drop shadow
}

\newcommand{\greencheck}{\textcolor{green}{\ding{51}}} 
\newcommand{\redx}{\textcolor{red}{\ding{55}}}

\title{Medical Triage as Pairwise Ranking: A Benchmark for Urgency in Patient Portal Messages}

\author{Joseph Gatto\textsuperscript{1}, Parker Seegmiller\textsuperscript{1}, Timothy Burdick\textsuperscript{2}, \\ \textbf{Philip Resnik\textsuperscript{3}, Roshnik Rahat\textsuperscript{1}, Sarah DeLozier\textsuperscript{2}, Sarah M. Preum\textsuperscript{1}} \\
\textsuperscript{1} Dartmouth College, Hanover NH\\
\textsuperscript{2} Dartmouth Health, Hanover NH\\
\textsuperscript{3} University of Maryland, College Park\\
\texttt{\{joseph.m.gatto.gr\}@dartmouth.edu}}

\begin{document}
\maketitle
\begin{abstract}
Medical triage is the task of allocating medical resources and prioritizing patients based on medical need. This paper introduces the first large-scale public dataset for studying medical triage in the context of asynchronous outpatient portal messages. Our novel task formulation views patient message triage as a pairwise inference problem, where we train LLMs to choose ``which message is more medically urgent" in a head-to-head tournament-style re-sort of a physician's inbox. Our novel benchmark PMR-Bench contains 1569 unique messages and 2,000+ high-quality test pairs for pairwise medical urgency assessment alongside a scalable training data generation pipeline.
PMR-Bench includes samples that contain both unstructured patient-written messages alongside real electronic health record (EHR) data, emulating a real-world medical triage scenario. 

We develop a novel automated data annotation strategy to provide LLMs with in-domain guidance on this task. The resulting data is used to train two model classes, \textbf{UrgentReward} and \textbf{UrgentSFT}, leveraging Bradley-Terry and next token prediction objective, respectively to perform pairwise urgency classification. We find that UrgentSFT achieves top performance on PMR-Bench, with UrgentReward showing distinct advantages in low-resource settings. For example, UrgentSFT-8B and UrgentReward-8B provide a 15- and 16-point boost, respectively, on inbox sorting metrics over off-the-shelf 8B models. Paper resources can be found at \url{https://tinyurl.com/Patient-Message-Triage}

\end{abstract}

\section{Introduction}

Recently, there has been great focus on integrating Large Language Models (LLMs) into clinical workflows \cite{liu2024survey, artsi2025large, tu2025towards, chen2024huatuogpt, hu2025systematic, wang-etal-2025-towards-adapting}. An emerging use case of LLMs in medicine has been their ability to help physicians manage and respond to the surging number of messages patients send their doctor through EHR-integrated web portals (also known as ``patient portals") \cite{holmgren2025trends, garcia2024artificial}. This surge has been correlated to increases in physician burnout \cite{apaydin2025secure, stillman2023death} and prior works have highlighted multiple directions in which NLP can be used to alleviate patient portal workloads via (i) patient message categorization \cite{harzand2023clinician}, (ii) LLM-drafted responses to patient messages \cite{nov2023putting, athavale2023potential, hu2025systematic, garcia2024artificial}, and (iii) eliciting missing information in patient messages through follow-up question generation \cite{gatto-etal-2025-follow}.  

\begin{figure}
    \centering
    \includegraphics[width=\columnwidth]{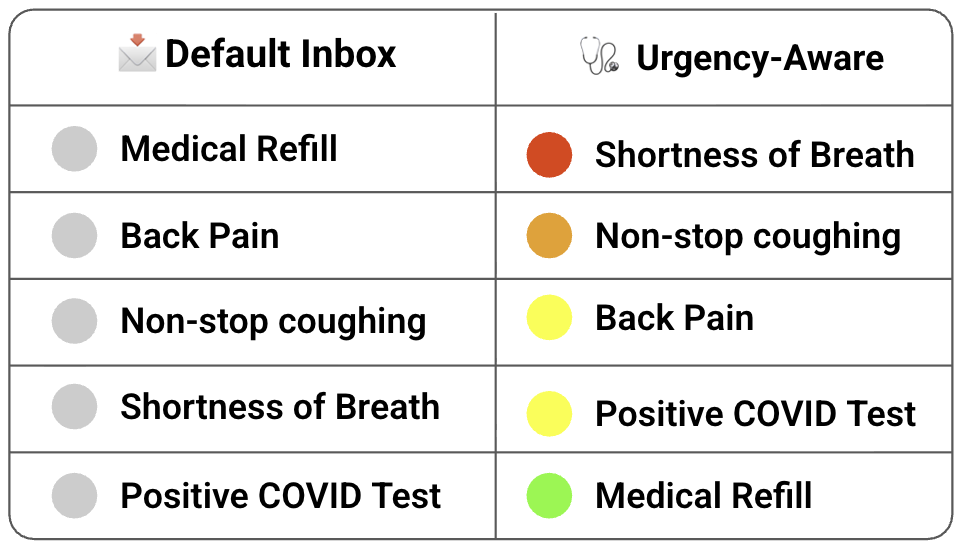}
    \caption{In this study, we introduce PMR-Bench, a novel dataset for evaluating LLM capacity to produce ``Urgency Aware" inboxes, where patient messages in clinicians' inboxes are sorted by medical urgency. Note that in a categorical setup, multiple messages can have a similar level of urgency.
    \label{fig:placeholder}} 
\end{figure}

An under-explored problem in this space is the task of \textbf{P}atient \textbf{M}essage \textbf{R}anking (PMR), whose goal is to help doctors prioritize/rank patient messages with higher degrees of medical urgency. PMR optimizes the ordering of messages in a doctor's inbox, directly influencing the patients that are responded to sooner \cite{apathy2024inbox}, leading to preventing care escalations for patients with more time-sensitive/urgent needs \cite{mermin2023use}. PMR is similar to prior works in patient message triage such as \cite{gatto2022identifying, harzand2023clinician, pmlr-v126-si20a, liu2025detecting} who solve the problem of portal message ranking by mapping messages into a discrete label space (e.g. \textit{urgent} vs \textit{non-urgent}). 

However, this paper aims to address the following \textbf{three limitations of triage-related prior work}. (i) Categorical definitions of urgency vary at a clinician- and organization-level, making prior work difficult to generalize to another setting. For example, clinicians of different specialties, training levels, years of experience, or geo-location (e.g., rural vs urban) usually have differing notions of what qualifies as an ``urgent" message \cite{quan2013perceptions}. The phenomenon of low inter-rater agreement can be found throughout clinical NLP \cite{wornow2024zeroshotclinicaltrialpatient, brake-schaaf-2024-comparing} and prior works have discussed this challenge as it pertains to medical triage \cite{naved2024contrasting}. (ii) Ranking messages based on discrete class labels provides only a weak ordering of messages, with no intra-class prioritization. (iii) Most prior work focuses on only messages specific to a single medical condition (COVID-19 \cite{gatto2022identifying, mermin2023use}), organ system (cardiology \cite{pmlr-v126-si20a}), or medical emergencies \cite{liu2025detecting}.

In this study, we address the gaps in prior work by posing message triage as a pairwise ranking problem instead of a classification problem. 

Specifically, we introduce a novel benchmark, \textbf{Patient Message Ranking (PMR)-Bench}, a pairwise text classification task covering a diverse array of medical conditions in primary care where the goal is to decide which of the two messages is \textit{more medically urgent}. Unlike classification, this task formulation is more directly connected to the real problem of deciding which messages should be treated as having higher priority. In addition, intuitively, a binary higher- versus lower-urgency comparison involves simpler comparison semantics than an ordinal set of three or more urgency categories. The ability to compute these comparisons inherently solves the ranking problem, as a PMR model can be deployed as the comparator in a sorting algorithm (e.g., bubble sort/quick sort) to rank a doctor's inbox based on medical urgency \cite{qin-etal-2024-large, zhuang2024setwise}.

PMR-Bench contains 1,569 unique patient messages with clinicians' ordinal annotations. This enables large-scale generation of data pairs for pairwise urgency detection (i.e. \textit{which of two patients is more medically urgent}) from two different medical communication platforms.  \textbf{First, PMR-Reddit} is a publicly available, curated set of patient messages from r/AskDocs --- an online forum where medical experts respond to patient queries. \textbf{Second}, \textbf{PMR-Synth} is a publicly available set of pairwise message comparisons using high-quality, synthetic patient portal messages, paired with real EHR data to emulate a real patient-portal environment in which urgency is determined using both patient message and structured EHR data. \textbf{Finally}, \textbf{PMR-Real} is a proprietary set of real-patient messages and corresponding EHR data sourced from a large regional hospital in the US. 

We explore two fine-tuning strategies for LLMs. The first is \textbf{UrgentSFT}, which uses Supevised Fine-Tuning (SFT) to adapt LLMs to our novel task. Furthermore, we introduce \textbf{UrgentReward}, which frames pairwise inference training in a reward modeling context. We show that UrgentReward 
achieves high performance with only an 8B parameter LLM. UrgentReward-8B outperforms GPT-OSS (120B) on this task and achieves 95\% of the performance of larger finetuned LLMs. We also define a set of task-specific metrics to evaluate model performance in this task. We summarize our contributions as follows: 

\begin{itemize}

    \item We introduce PMR-Bench, the first large-scale benchmark for pairwise medical urgency assessment. PMR-Bench includes patient messages paired with real structured EHR data --- emulating a realistic patient message triage environment. We benchmark 8 LLMs on our novel task. We will make our dataset available on Huggingface. 
    
    \item We develop two models,  UrgentReward and UrgentSFT, which are pairwise inference approaches for determining which of two patient messages should be attended to first. Our methods optimize for accuracy and efficiency with strong performance across 4B, 8B, 27B, and 32B parameter model variants --- making our methods more suitable for low-resource settings. For example, UrgentSFT-8B and UrgentReward-8B provide a 15- and 16-point boost, respectively, on inbox sorting metrics over off-the-shelf 8B models.

\end{itemize}
\section{Related Work}

\paragraph{LLMs for Document Ranking: } 
Document ranking is a common Information Retrieval task that aims to rank document relevance to a search query \cite{xu2025surveymodelarchitecturesinformation}. In recent years, it has become common to employ computationally expensive document ranking models, including LLMs, on small lists of documents, refining the outputs of more efficient methods which have been applied to larger document sets \cite{robertson2009probabilistic}. For example, \cite{sun-etal-2023-chatgpt, qin-etal-2024-large, zhuang2024setwise} introduce state-of-the-art re-ranking methods in the context of LLMs, with recent works \cite{zhuang2025rank} using reasoning models for document ranking. 

Given a model that can perfectly determine which of two documents is most relevant to a query, we can leverage the theoretical guarantees of sorting algorithms to re-rank a set of documents using pairwise comparisons \cite{zhuang2024setwise}. In practice, LLMs may make mistakes and thus become sensitive to the initial ordering of the documents. One can avoid order sensitivity by computing all $n \choose 2$ comparisons and sorting by win rate. This method is highly accurate, but limited by higher inference cost when compared to more efficient approaches such as pointwise or listwise re-ranking \cite{qin-etal-2024-large, zhuang2024setwise}. 

In this study, we focus exclusively on pairwise re-ranking strategies when sorting patient portal messages. This design choice is motivated by the following three task-specific constraints. (i) PMR is a safety-critical task, demanding that we choose a sorting method that provides stronger guarantees and is less sensitive to the initial document order. (ii) PMR can afford higher latency, e.g., a clinician's inbox can be re-sorted during the hours the clinician is away / seeing patients. Also, unlike other IR tasks such as web page ranking, where users only visit a handful of search results, in PMR, clinicians eventually attend to all patient messages within a time window (e.g., 2-3 business days). (iii) In production, a PMR system can reuse past comparisons, precluding the need to re-sort the full inbox with $n \choose 2$ comparisons every time a new patient message arrives.

\paragraph{Patient Messages Triaging}

Prior works have explored employing NLP methods to classify patient messages based on their urgency. For example, \cite{pmlr-v126-si20a} employ transformer-based classifiers to categorize patient messages into [\textit{urgent, medium, non-urgent}] categories. Similarly,  \cite{harzand2023clinician} study how to route patient messages using five categories: [\textit{urgent, clinician, prescription refill, schedule, form}].  \citet{10.1093/jamiaopen/ooae078} study the capacity of a BERT-based model \cite{devlin-etal-2019-bert} to flag patient messages with high acuity. \citet{mermin2023use} use NLP to automatically detect patients with COVID while \citet{gatto2022identifying} use BERT-based methods to detect the severity of messages from online medical Q\&A forums. Other related works, such as \cite{lu-etal-2024-triageagent} have used LLMs to determine the urgency of synthetic clinical vignettes from the Emergency Department (ED) into 5 categorical rankings determined by the Emergency Severity Index (ESI).\footnote{Although relevant, ESI rankings do not apply to primary care / outpatient clinics due to significant differences in medical content (e.g. portal messages have far-fewer critical emergencies).} In this study, we extend prior work by developing an urgency scale that extends to domains such as primary care while re-framing the urgency classification task as a pairwise inference problem for enhanced accuracy and generalizability.

Additionally, different NLP approaches, and more recently LLMs, have been applied broadly to patient messages ranging from writing responses to patient messages \cite{nov2023putting, athavale2023potential, garcia2024artificial} to asking patients follow-up questions to elicit missing information \citet{gatto2025followupquestiongenerationenhanced}. 
\section{Methods}

\begin{figure}
    \centering
    \includegraphics[width=\columnwidth]{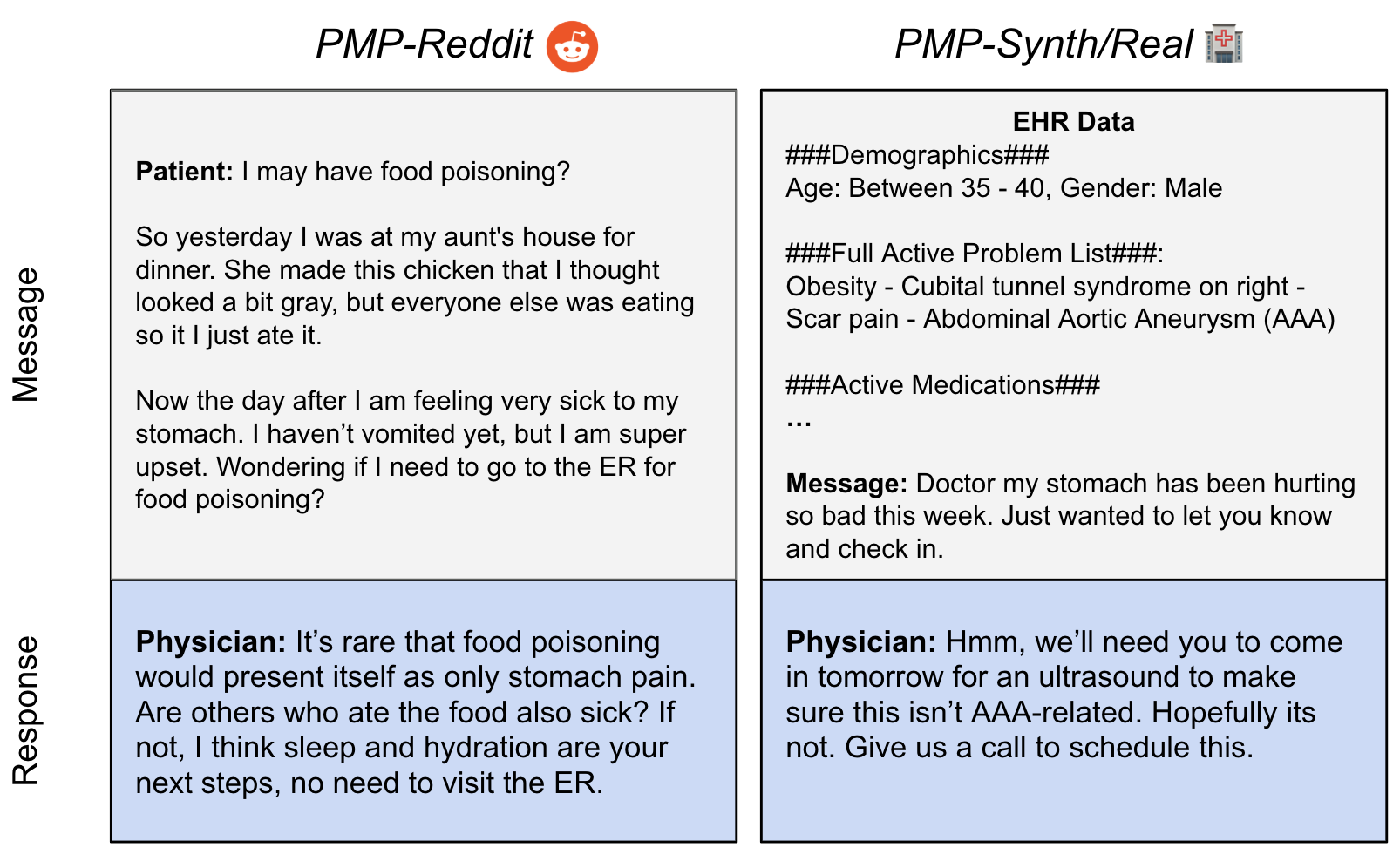}
    \caption{Example data format for PMR-Reddit and PMR-Synth/Real. Data from Reddit is unstructured text, with responses from moderator-verified clinical experts. Data from PMR-Synth and PMR-Real include linked EHR data, as the EHR data impact triage decisions. }
    \label{fig:data_example_v2}
\end{figure}

\subsection{PMR as a Pairwise Ranking Task } \label{sec:task_formulation}

Just as we are biased towards reading emails in the order they appear in our inbox, physicians are biased toward reviewing patient messages/queries in the order they are presented \cite{apathy2024inbox}. Thus, for a set of patient messages $P = \{p_1, \dots, p_n\}$, the goal of PMR is to learn the optimal ordering $P'$ such that messages with a higher degree of medical urgency are ranked higher in $P'$. The resulting sort enables patients with greater medical needs to receive clinicians' attention sooner. 

Each $p_i \in P$ may contain an associated structured EHR record $e_i$. EHR records contain relevant patient details such as age, gender, medication list, diagnosis history, and active problem list. Such context is often taken into consideration when patient messages are reviewed \cite{ozkaynak2014examining} --- making PMR a multi-modal inference problem. 

\begin{table}[!t]
    \centering
    \small
    \begin{tabular}{l|ccc}
        \toprule
                       & Reddit & Synth & Real \\
        \midrule
     Avg Tokens & 228 $\pm 153$  & 457 $\pm 100$  & 511 $\pm 147$ \\
     Unique Msg & 1121             & 60 & 388 \\
        \midrule        
        Level 1     & 126                 & 10 & 34 \\
        Level 2     & 45                  & 10 & 68 \\
        Level 3     & 59                  & 10 & 79 \\
        Level 4     & 384                 & 10 & 100 \\
        Level 5     & 283                 & 10 & 100 \\
        Level 6     & 324                 & 10 & 7 \\
        \midrule  
        Has EHR        & \redx & \greencheck & \greencheck \\
        \bottomrule
    \end{tabular}
    \caption{PMR-Bench dataset statistics. Token numbers estimated using the Qwen3 tokenizer.}
    \label{tab:data-stats}
\end{table}

Following \cite{qin-etal-2024-large}, we sort $P$ via pairwise comparisons across patients. For example, for any two patient messages $(p_i, p_j)$, we can feed them to a model $f$ whose job is to determine which of the two patient messages should be attended to first. We can then re-sort a clinician's inbox using $f$ as the comparator in a sorting algorithm or to compute all $n \choose 2$ comparisons and sort messages based on their "win-rate" \cite{shah2018simple}. In this study, we focus on the later, as $|P|$ is often small and sorting by win-rate gives a better upper-bound on performance as methods are often sensitive to the initial ordering of samples. 

\subsection{PMR-Bench Dataset Overview}
In Table \ref{tab:data-stats} we provide an overview of PMR-Bench, which contains 1,569 unique healthcare messages from multiple data sources. (i) \textbf{PMR-Reddit} contains messages sourced from the subreddit r/AskDocs, (ii) \textbf{PMR-Real} leverages a proprietary corpus of patient messages from a large regional hospital in the US, and (iii) \textbf{PMR-Synth} uses expert-written messages which aim to emulate the style and prose of PMR-Real while enabling us to share high-quality data publicly for reproducibility. Example data from each source is shown in Figure \ref{fig:data_example_v2}. Due to the high cost and challenges of collecting reliable annotation at scale, we developed a reproducible annotation method that infers labels from clinician responses to messages that are readily available in real responses to patient messages on r/AskDocs and from real physicians in PMR-Real (more details in Section \ref{subsec:annotation}). For PMR-Synth, we had clinical experts directly annotate pairs of messages. 

In the remainder of this section, we first describe in detail our automated urgency annotation process followed by additional details for each dataset. 

\subsubsection{Urgency Annotation}
\label{subsec:annotation}
We curate pairwise urgency labels by leveraging existing expert responses to patient messages. To accomplish this, we aim to classify each response into an ordinal set of urgency categories. To the best of our knowledge, there does not exist any accepted standard for medical urgency classification outside of emergency medicine \cite{tanabe2004emergency} --- which has limited applicability to patient portal messages. We have thus developed a 6-tier ordinal scale for labeling urgency in patient portal messages in collaboration with a team of clinicians at a large regional medical center. Our scale goes from Level 1 (Most Urgent --- Emergency Attention Needed) to Level 6 (Least Urgent --- No Medical Attention Needed). The full label space with examples is shown in Table \ref{tab:label_definitions} in the Appendix. 

Now, for a sample $(q,r)$ where $q$ is a patient message and $r$ is a response from a clinical expert, we use an LLM $g$ to classify the response $r$ into the above scale to determine the degree of urgency of the message.\footnote{We used GPT-5 for public datasets. For sensitive data, we used GPT-OSS 20B on our secure computing server.} For example, patients instructed to go to the ED are classified as ``Level 1" while patients given self-care strategies are classified as "Level 5". We then create pairwise annotation for two messages $(q_i, q_j)$ based on the relative ranking of their respective responses $(g(r_i), g(r_j))$. In Appendix \ref{apn:labeling_details} we demonstrate each of the six levels of urgency, and outline additional steps taken to ensure reliability and quality of annotation (e.g. sample quality filtering and judge models to validate the label accuracy of each pair of messages).

We apply the above strategy to create PMR-Reddit and PMR-Real. As discussed in Section \ref{sec:task_formulation}, samples in PMR-Real each have an associated EHR record, while PMR-Reddit samples do not. To create a publicly available multi-modal version of the task, we also construct PMR-Synth. As synthetic patient messages will not have a corresponding expert response to infer the urgency label like the other two datasets, we had a team of clinicians (i.e., triage nurses and physicians) provide pairwise annotation directly to pairs of synthetic patient messages. Each patient message in PMR-Synth is paired with a de-identified EHR record from a real patient at our partner hospital and is made publicly available for reproducibility. We provide sample data as a supplement to this submission for reference. More details for PMR-Synth annotation can be found in Appendix \ref{sec:pmr_synth}. 

We note that the design choice to go from an \textbf{ordinal urgency scale to a pair-wise urgency comparison} has many advantages. First, having a quantitative measure of gap in urgency between two samples allows us to define the difficulty of a pair, where samples far apart on the scale should be easier to classify compared to samples which are close-by. In Section \ref{sec:results} we show that all model performances go down as difficulty goes up, validating the quality of our annotations. Furthermore, having a label for a single message allows us to run an experiment where we test model capacity to predict the label directly.

\subsubsection{Dataset Details}\label{sec:data_deets}

\noindent \textbf{PMR-Reddit:} We source messages from r/AskDocs, a forum where patients can request feedback from verified clinical experts on Reddit. Using an LLM, we filter for those patients who are looking for feedback on acutely onset symptoms. We then classify each comment from a verified expert clinician using GPT-5. Note that we only consider posts with expert comments which (i) received at least 5 upvotes, or (ii) had comment sections with agreement across multiple expert comments. We then classify each comment into our 6-level urgency scale (see Table \ref{tab:label_definitions}). Our resulting dataset has 1,121 unique posts with the label distribution shown in Table \ref{sec:data_deets}. The ordinal label assigned to the expert comment is used to sample posts for pairwise comparisons. The PMR-Reddit test set has a total of 1502 pairwise comparisons derived from 362 posts. 

\noindent \textbf{PMR-Real:} We source a set of real patient portal messages from a large regional medical center in the United States. Similar to the preprocessing of PMR-Reddit, we filter for patient messages describing acutely onset symptoms and classify the responses to the resulting messages into our ordinal urgency scale. A key difference between PMR-Real and PMR-Reddit is the inclusion of structured EHR data. Each PMR-Real message has as its linked EHR data, including the patient's (i) active problem list, (ii) recent diagnoses, (iii) active medications, and (iv) demographic information. Note that we are unable to publicly release PMR-Real due to the sensitive nature of the dataset.

\noindent \textbf{PMR-Synth:} Messages in PMR-Synth were written and curated by expert members of the study team and subsequently reviewed by additional clinical experts to ensure high-quality, realistic samples covering a wide array of medical topics. Note, we intentionally opted to hand-craft each message as LLM-generated content struggled to match realistic patient tone and style. The EHR record associated with each message is the de-identified EHR of a real patient from the same pool of patients used to create PMR-Real. Unlike PMR-Reddit and PMR-Synth, we had a team of medical experts directly classify all $n \choose 2$ message pairs for two separate inboxes of size 30, annotating which of two patients should receive priority medical care. We treat one inbox as a training and the other as a testing set. We report inter-annotator agreement for PMR-Synth in Table \ref{tab:iaa_metrics} with additional annotation details in Appendix \ref{sec:pmr_synth}.

\noindent \textbf{Sample Difficulty Quantification: } Each message now has a label 1-6 describing how urgent the message is. We consider the difference between the two labels to be a proxy for pairwise sample difficulty. \textbf{Easy:} Samples with a difficulty level of at least 4 (e.g. Level 1 vs Level 5/6). \textbf{Medium: } Samples which are 2-3 difficulty levels apart (e.g. Level 1 vs Level 3/4). \textbf{Hard: } Samples which are less than 2 difficulty levels apart (e.g. Level 2 and Level 3). It should be noted that this was an empirical design choice and can be explored further in future work.

\subsection{UrgentSFT} \label{results: classify}

To solve the pairwise urgency task, we develop UrgentSFT, a two-step inference procedure for classifying pairwise medical urgency. Consider an LLM $f$ which processes a pair of texts $f(a,b)$ where $a$ and $b$ denote two different patient messages (our UrgentSFT prompt can be found in Figure \ref{prompt:urgent_sft}). UrgentSFT essentially asks the model to output a probability (via the `YES' token as in other IR tasks \cite{liang2023holisticevaluationlanguagemodels}) deciding if message $b$ should be attended to before message $a$ based on their respective medical urgency. For a given pair of patient messages $(a,b)$ where $b$ was annotated as the more urgent message, we evaluate model correctness using the following probability difference 
\[\eta = P(\mathrm{YES} | f(a,b)) - P(\mathrm{YES} | f(b,a))\]
If $\eta$ > 0, then the model is correct, as it has successfully attributed a higher probability to the more urgent patient message. We formulate UrgentSFT as a two-step inference leveraging probability scores to help prevent ties and improve models' sensitivity to input order. Using patient messages which have been classified into our ordinal urgency scale, we create training pairs for UrgentSFT fine-tuning.

\subsection{UrgentReward}
UrgentReward frames pairwise classification using a Bradley-Terry (BT) loss \cite{bradley1952rank} to help the LLM better internalize a relative urgency ranking among patient messages. We note that this is in contrast to UrgentSFT which is trained using the standard next token prediction objective. 

For a given patient message $t$, we create UrgentReward training triplets ($t$, $t_m$, $t_l$) where $t_m$ and $t_l$ are patient messages that are more and less urgent than $t$, respectively. Let $t_p$ denote a prompt which contains within it the message $t$. Specifically, the UrgentReward prompt template (shown in Figure \ref{prompt:urgent_reward}) asks the model to \textit{write} a message more medically urgent than the one provided. We thus frame the task of quantifying how much more urgent one message is when compared to another as the scoring of a prompt completion --- aligning our work with prior studies on reward modeling. We fine-tune a reward model using $U_r$ using the TRL package \cite{vonwerra2022trl}, which uses a Bradley-Terry objective to maximize $ U_r(t_p, t_m) - U_r(t_p, t_l)$.

At test time, we apply UrgentReward similarly to UrgentSFT. For a given pair of patients $(a, b)$, we run two inferences $f(a, b) = s_1$ and $f(b, a) = s_2$. If $s_1 > s_2$, this means that $b$ is more urgent than $a$. It is crucial to note that the \textit{pairwise} application of the BT model makes it distinct from prior works in IR that apply BT models in a pointwise re-rank setting. We perform two BT inferences per pair, and use those scores only to produce a pairwise classification label. 

To leverage existing knowledge on scoring prompt completions, all UrgentReward models are fine-tuned on Qwen-based SkyWork-Reward-v2 models \cite{liu2025skywork}, which are state-of-the-art LLM-based sequence classifiers pre-trained on 26 million preference pairs.

\begin{table*}[!ht]
\centering
\small
\begin{tabular}{lccccccccccccc}
\toprule
& \multicolumn{4}{c}{PMR-Reddit} & \multicolumn{4}{c}{PMR-Synth} & \multicolumn{4}{c}{PMR-Real} \\
\cmidrule(lr){2-5} \cmidrule(lr){6-9} \cmidrule(lr){10-13}
   & Easy & Med & Hard & Total 
        & Easy & Med & Hard & Total
        & Easy  & Med & Hard & Total \\
\midrule 
\textit{Number of Test Pairs} & 318 & 736 & 448 & 1502 & 75 & 175 & 185 & 435 & 51 & 274 & 241 & 566\\
\midrule 
\textit{Instruct Models} \\
Qwen3-4B                                      & 0.85 & 0.72 & 0.66 & 0.73 & 0.76 & 0.66 & 0.50 & 0.61 &  0.86 & 0.74 & 0.51 & 0.66 \\
Qwen3-8B                                      & 0.85 & 0.73 & 0.68 & 0.74 & 0.80 & 0.74 & 0.55 & 0.67 & 0.88 & 0.72 & 0.60 & 0.68\\
Qwen3-32B                                    & 0.89 & 0.76 & 0.69 & 0.77 & 0.77 & 0.70 & 0.51 & 0.63 & 0.90 & 0.76 & 0.58 & 0.70 \\
MedGemma-27b                                 & 0.89 & 0.74 & 0.70 & 0.76 & \textbf{0.95} & 0.73 & \textbf{0.63} & \underline{0.72} & 0.80 & 0.74 & 0.60 & 0.68 \\
\midrule 
\textit{Deep Reasoning Models} \\
Qwen3-32B-R                                  & 0.81 & 0.64 & 0.56 & 0.65 & 0.68 & 0.58 & 0.38 & 0.51 & 0.78 & 0.58 & 0.41 & 0.53 \\
GPT-OSS*                                     & 0.93 & 0.76 & 0.67 & 0.77 & 0.79 & 0.60 & 0.50 & 0.59 & 0.86 & 0.64 & 0.56 & 0.63 \\
\midrule
\textit{UrgentSFT} \\
Qwen3-4B$_{SFT}$                                 & 0.84 & 0.69 & 0.63 & 0.70 & 0.81 & \underline{0.75} & 0.50 & 0.65 & \underline{0.90} & 0.79 & \underline{0.67} & \underline{0.75} \\
Qwen3-8B$_{SFT}$                                 & 0.90 & 0.75 & 0.69 & 0.76 & 0.83 & 0.68 & 0.52 & 0.64 & \textbf{0.92} & 0.80 & 0.64 & 0.74 \\ 
Qwen3-32B$_{SFT}$                                & \underline{0.96} & \textbf{0.86} & 0.84 & \underline{0.87} & 0.92 & 0.70 & 0.60 & 0.69 & $\dagger$ & $\dagger$ & $\dagger$ & $\dagger$ \\ 
MedGemma-27B$_{SFT}$                    & \textbf{0.98} & \underline{0.85} & \textbf{0.87} & \textbf{0.88} & \underline{0.93} & \textbf{0.77} & 0.60 & \textbf{0.73} & \textbf{0.92} & 0.78 & \textbf{0.72} & \textbf{0.77} \\ 
\midrule
\textit{UrgentReward} \\
Reward-4B$_{base}$     & 0.79 & 0.67 & 0.55 & 0.66 & 0.75 & 0.64 & 0.53 & 0.61 & 0.82 & 0.75 & 0.61 & 0.70 \\
Reward-4B$_{urgent}$      & 0.91 & 0.80 & 0.80 & 0.82 & 0.80 & 0.66 & 0.56 & 0.64 & \textbf{0.92} & \textbf{0.82} & \underline{0.63} & \underline{0.75} \\ 
Reward-8B$_{base}$     & 0.80 & 0.67 & 0.55 & 0.66 & 0.84 & \underline{0.75} & 0.54 & 0.68 & 0.86 & 0.72 & 0.57 & 0.67 \\
Reward-8B$_{urgent}$    & 0.93 & 0.82 & \underline{0.85} & 0.85 & 0.91 & 0.73 & \underline{0.62} & 0.71 & \textbf{0.92} & \underline{0.81} & \underline{0.63} & 0.74 \\
\bottomrule
\end{tabular}
\caption{Pairwise classification accuracy on each dataset, reported by each difficulty level (easy, medium (med), hard). Top performing models are \textbf{bold}. 2nd highest performing models are \underline{underlined}. We find that UrgentSFT-MedGemma shows the highest performance with UrgentReward showing comparable scores with a much smaller model. * GPT-OSS experiments use the 120B model for PMR-Reddit / Synth and 20B for real. $\dagger$ We do not have the computational capacity to fine-tune Qwen3-32B on our secure server. }
\label{tab:pairwise-results}
\end{table*}

\section{Experimental Setup}
In this section, we describe the baselines and metrics used to evaluate our methods in this study. We consider two evaluation settings. We first consider an \textbf{intrinsic evaluation} where we directly assess the binary classification accuracy of each model on PMR-Bench pairs. We then conduct an \textbf{extrinsic} evaluation, directly assessing how well each model can sort a clinician's inbox.

\subsection{Intrinsic Evaluation}
Following \cite{lambert2025rewardbench} we report the classification accuracy in Table \ref{tab:pairwise-results} as our evaluation setup is structurally similar to that of a reward modeling evaluation. We report overall accuracy as well as per-difficulty accuracy, where \textit{difficulty} is defined by the difference in ordinal triage rankings as defined in Section \ref{sec:data_deets}. 

\subsection{Extrinsic Evaluation}\label{sec:extrinsic} 

\paragraph{Data Preparation: } We sample messages of varying urgency levels to create a diverse inbox of approximately 30 messages per corpus. This is motivated by \cite{adler2020electronic}, who found that clinicians received 229 messages per week, which is approximately 32 messages per day. Please see Appendix Table \ref{sec:extrinsic_eval} for the urgency level distribution of each inbox.

\noindent \textbf{Inference: } For an inbox with $n$ messages in it, we compute $n \choose 2$ pairwise comparisons. Each time a sample is deemed more urgent than another, we increment it's score by (1 + $
\eta$), where $\eta$ is the difference in normalized probabilities / reward scores. The inbox is then sorted based on the total score of each sample. 

\noindent \textbf{Evaluation Metrics:} We convert the urgency labels from our ordinal urgency scale into relevancy scores to map our problem into an information retrieval setting. E.g., Level 1 samples have the highest relevancy scores as these patient messages should be attended to before \textit{lower urgency} patient messages. We can then compute classic IR metrics such as NDCG@K \cite{DBLP:journals/tois/JarvelinK02} directly from our relevancy-mapped samples. One drawback of NDCG is it will not properly penalize the model for sorting a highly urgent message to the bottom of the inbox which is a safety-critical error in this task. Inspired by prior works with similar motivations to address content sorted at the bottom of a list \cite{10.1145/3340531.3412123, crowdcent_symmetric_ndcg_2025} we report a tail-normalized NDCG (T-NDCG) which penalizes a model for sorting urgent information to the bottom. Specifically, for an inbox $I$ which has been sorted by a ranking model, the T-NDCG@k is $NDCG@K(I) - NDCG@K(r(I))$, where $r(I)$ is the reverse sorting of $I$ --- penalizing the model for placing urgent messages at the bottom of the list. We use the \textit{ranx} \cite{ranx} package to compute IR metrics in this study.

\subsection{Baseline Models}
\textbf{Instruct Models: } We explore four non-reasoning models, Qwen3-4/8/32B  \cite{qwen3technicalreport} with thinking disabled, and a medical LLM, Medgemma-27b-text-it \cite{sellergren2025medgemmatechnicalreport}. \\
\textbf{Reasoning Models: } We explore two reasoning models, Qwen3-32B and GPT-OSS \cite{openai2025gptoss120bgptoss20bmodel}. Unlike instruct models, which use the probability of the ``YES" token, we attribute a probability of 1.0 when the model predicts ``YES" as the reasoning process makes use of token probabilities less meaningful.

\noindent \textbf{Training Data: } The training data for UrgentSFT and UrgentReward are exactly the same for each of the three datasets. We curate training triplets from the pool of samples we classified into our 6-label scale. For UrgentReward, this translates into an anchor sample and then a \textit{chosen} and \textit{rejected} completion used for training e.g. (Anchor, More Urgent, Less Urgent). The same triplet is converted into multiple SFT samples (e.g.  (Anchor, More Urgent, YES) and (Anchor, Less Urgent, NO)). Additional training details, including training data, data distributions, fine-tuning strategy, and model selection is shown in Appendix \ref{apn:training_deets}.

\noindent \textbf{Multi-Class Baseline: } For our extrinsic evaluation, we also evaluate LLM capacity to predict the class label directly. We explore GPT-OSS as well as MedGemma-27B and Qwen3-32B with and without SFT on this task. As multi-class models will produce rankings with many ties, we report an expected T-NDCG \cite{mcsherry2008computing} in Table \ref{tab:sorting_results_table}, which in our implementation is the average T-NDCG given numerous intra-class shuffles.

\section{Results}\label{sec:results}
\subsection{Intrinsic Evaluation: Pairwise Classification}

Table \ref{tab:pairwise-results} displays our pairwise classification results. We find that on PMR-Reddit and PMR-Synth, UrgentSFT with MedGemma-27b achieves the highest overall performance. In general, PMR-Reddit scores are higher than PMR-Synth and PMR-Real. This result is intuitive as models do not need to process structured EHR information in PMR-Reddit. Also noteworthy is that PMR-Reddit has a much larger training set, likely contributing to higher performance. However, our ablation study in Table \ref{tab:ablation_reddit_data_size} shows that UrgentReward has the capacity to get more out of less training data when applied to smaller models, making it a viable option when fewer resources are accessible.

Noteworthy is the comparison between baseline instruct models and reasoning models. We find that Qwen3-32B without reasoning out-performs Qwen3-32B with reasoning. We believe this may be due to input order biases being exaggerated by reasoning as well as the Instruct Models having the advantage of tie-breaking via token probabilities. 
Overall, our methods substantially reduce pairwise triage error on all real datasets, demonstrating that UrgentSFT and UrgentReward can deliver meaningful real-world triage improvements while remaining lightweight and deployable with smaller language models

\begin{table}[t]
\centering
\small
\setlength{\tabcolsep}{3pt} 
\begin{tabular}{lcccccc}
\toprule
& \multicolumn{2}{c}{Reddit} & \multicolumn{2}{c}{Synth} & \multicolumn{2}{c}{Real} \\
\cmidrule(lr){2-3}\cmidrule(lr){4-5}\cmidrule(lr){6-7}
 & @10 & @30 & @10 & @30 & @10 & @30 \\
\midrule
\textit{Multi-Class} &&&&&& \\
MedGem27B                & 0.49 & 0.25 & 0.54 & 0.24 & 0.64 & 0.32 \\
MedGem27B$_{SFT}$        & 0.62 & \underline{0.32} & 0.27 & 0.18 & 0.66 & 0.35 \\
Qwen32B                     & 0.59 & 0.28 & 0.50 & 0.26 & 0.62 & 0.34 \\
Qwen32B$_{SFT}$             & 0.50 & 0.24 & 0.57 & 0.27 & $\dagger$      & $\dagger$ \\
GPT-OSS                      & 0.40 & 0.18 & 0.48 & 0.23 & 0.48 & 0.23 \\
\midrule
\textit{0-Shot Pairs} &&&&&& \\
Qwen8B                        & 0.41 & 0.20 & 0.52 & 0.27 & 0.52 & 0.18 \\
Qwen32B                      & 0.54 & 0.25 & 0.62 & 0.28 & 0.42 & 0.17 \\
MedGem27B               & 0.38 & 0.16 & \underline{0.66}& 0.32 & 0.70 & 0.31 \\
GPT-OSS*                        & 0.44 & 0.24 & 0.58 & 0.28 & 0.59 & 0.32 \\
\midrule
\textit{UrgentSFT Pairs} &&&&&& \\
Qwen8B$_{SFT}$            & 0.51 & 0.23 & 0.62 & 0.30 & \textbf{0.77} & \textbf{0.39} \\
Qwen32B$_{SFT}$          & \textbf{0.77} & \textbf{0.36} & \textbf{0.73}& \underline{0.34}& $\dagger$ & $\dagger$ \\
MedGem27B$_{SFT}$    & 0.61 & 0.29 & \underline{0.66}& \textbf{0.35}& \underline{0.75} & \textbf{0.39} \\
\midrule  
\textit{UrgentReward Pairs} &&&&&& \\
Reward-4B$_{Base}$      & 0.22 & 0.10 & 0.37 & 0.18 & 0.48 & 0.18 \\
Reward-4B$_{Urgent}$  & \underline{0.69} & 0.30 & 0.55 & 0.26 & 0.65 & 0.37 \\
Reward-8B$_{Base}$      & 0.33 & 0.14 & 0.57 & 0.26 & 0.54 & 0.20 \\
Reward-8B$_{Urgent}$  & 0.58 & 0.26 & 0.64 & 0.29 & 0.71 & \underline{0.38} \\
\bottomrule
\end{tabular}
\caption{Results of our extrinsic evaluation where each model is tasked with re-ranking a clinician's inbox. We report the T-NDCG metric (as described in section \ref{sec:extrinsic}) at k = 10 and k = 30. *GPT-OSS 120B used for Reddit/Synth and 20B used for Real due to resource constraints. $\dagger$ Experiment cannot be run due to resource constraints.}
\label{tab:sorting_results_table}
\end{table}

\subsection{Extrinsic Evaluation: Inbox Sorting}

The results of extrinsic evaluation are presented in Table \ref{tab:sorting_results_table}. The T-NDCG@30 metric considers the full inbox, reflecting broad-scale sorting quality for a given model. In contrast, T-NDCG@10 metrics emphasizes the top and bottom of the list, more-heavily rewarding correct placement—and penalizing misplacement—of highly urgent messages. The larger performance gap at @10 suggests our models are particularly effective at handling more urgent samples. On \textbf{PMR-Reddit} we find that UrgentSFT with Qwen3-32B is our highest performing model. Notably, this model outperforms the multi-class baseline, supporting our hypothesis that pairwise inference is more effective. On \textbf{PMR-Synth} we similarly find that our top-performing model is UrgentSFT with Qwen3-32B. Finally, on \textbf{PMR-Real} we see that UrgentSFT-8B is the top-performing models, with a T-NDCG of 0.77 and 0.39, @ 10 and @ 30, respectively.

Also, while some multi-class baselines appear competitive, their T-NDCG scores can vary greatly given different shuffles of the discrete class labels generated by the model, as demonstrated by high standard deviations of these models (in Appendix \ref{apn:multi_class_instability}). This makes pairwise inference not only more effective, but a more stable ranking mechanism. 

Due to space constraints, we refer the reader to the Appendix for additional ablation experiments and discussions. For example, in Appendix \ref{sec:training_size_ablation} we show how performance changes with varying training set sizes, with UrgentReward showing strong sample efficiency. In Appendix \ref{sec:ehr_ablation} we analyze PMR-Synth performance with and without EHR data. Finally, Appendix \ref{apn:error_analysis_bias} performs a brief analysis of model biases on patient demographics, demonstrating how model performance varies based on the age and gender of a patient.

\section{Conclusion}

In this study, we re-define patient portal message triage via a novel benchmark PMR-Bench, which frames triage as a pairwise inference problem. Our data leverages expert annotation and is first-of-its-kind to include structured EHR data alongside patient-written queries for medical triage.  Our results demonstrate that, on average, our two models, UrgentReward and UrgentSFT improve ranking performance over all baseline approaches, producing SOTA inbox sorting models. 
\section{Limitations}

While this work takes a strong step towards solving the issue of pairwise urgency classification, other practical considerations must be taken before deployment. For example, one may want to avoid the case where a low-urgency message is continually placed on the bottom of the inbox, causing longer-than-usual response delays. A real-world system may want to include time-in-inbox as a factor to abide by policies that pertain to response times (e.g. some healthcare systems require clinicians to respond within 72 hours regardless of message urgency). 

Our PMR-Real results are limited in that we are restricted to experimentation on a secure machine that cannot access the internet and only has a single 40GB GPU for experimentation. This makes benchmarking of proprietary LLMs and larger open-source models infeasible. Furthermore, we are unable to release the PMR-Real dataset due to institutional IRB policies. We release PMR-Synth dataset to mitigate this issue, which aims to mimic the style and format of samples in PMR-Real. 

Outside the scope of this submission was an in-depth analysis of model biases to different demographic and/or medical background traits. It may be the case that LLMs over or under triage certain subpopulations and this should be investigated before deployment.

Finally, one data limitation is that while a subset of PMR-Bench has gone under expert review, some labels are extracted from expert responses using LLMs. While ideally all samples could have undergone human review, we believe the classification of expert responses alongside extensive postprocessing (as discussed in Appendix \ref{apn:data_quality}) ensures the reliability and quality of the data while permitting us to study this problem at a greater scale. This particularly applies to our test sets, which underwent multiple rounds of filtering to capture high-quality triage data, i.e., it was clear from the clinician's response that a given patient was more urgent than another. 

\section{Ethical Considerations}
This study was conducted under IRB approval from the submitting author's institution. All publicly released EHR data has been de-identified and approved for release. All processing of any sensitive patient information was performed on a secure computing server with no internet access, hosted by the submitting authors institution. The PMR-Reddit samples are IRB-exempt and can be shared following Reddit's data usage policy. 

We further wish to highlight that while this work aims to sort patient messages by their medical urgency, it is generally the case that \textit{all messages are addressed/processed by clinicians}. All patient messages must be responded to as all users are deserving of the medical attention they are requesting. In this study, we aim to address preventable care escalation, where patients with more urgent issues are not addressed in a timely manner, which can lead to care escalations, e.g., hospitalization, admission to emergency room, delayed care, worsening medical symptoms, and other care-related inefficiencies. 

\bibliography{Triage}

@article{garcia2024artificial,
  title={Artificial intelligence--generated draft replies to patient inbox messages},
  author={Garcia, Patricia and Ma, Stephen P and Shah, Shreya and Smith, Margaret and Jeong, Yejin and Devon-Sand, Anna and Tai-Seale, Ming and Takazawa, Kevin and Clutter, Danyelle and Vogt, Kyle and others},
  journal={JAMA Network Open},
  volume={7},
  number={3},
  pages={e243201--e243201},
  year={2024},
  publisher={American Medical Association}
}

@article{hu2025systematic,
  title={A systematic review of early evidence on generative AI for drafting responses to patient messages},
  author={Hu, Di and Guo, Yawen and Zhou, Yiliang and Flores, Lidia and Zheng, Kai},
  journal={npj Health Systems},
  volume={2},
  number={1},
  pages={27},
  year={2025},
  publisher={Nature Publishing Group UK London}
}

@article{quan2013perceptions,
  title={Perceptions of urgency: defining the gap between what physicians and nurses perceive to be an urgent issue},
  author={Quan, Sherman D and Morra, Dante and Lau, Francis Y and Coke, W and Wong, Brian M and Wu, Robert C and Rossos, Peter G},
  journal={International Journal of Medical Informatics},
  volume={82},
  number={5},
  pages={378--386},
  year={2013},
  publisher={Elsevier}
}

@article{robertson2009probabilistic,
  title={The probabilistic relevance framework: BM25 and beyond},
  author={Robertson, Stephen and Zaragoza, Hugo and others},
  journal={Foundations and Trends{\textregistered} in Information Retrieval},
  volume={3},
  number={4},
  pages={333--389},
  year={2009},
  publisher={Now Publishers, Inc.}
}

@inproceedings{qin-etal-2024-large,
    title = "Large Language Models are Effective Text Rankers with Pairwise Ranking Prompting",
    author = "Qin, Zhen  and
      Jagerman, Rolf  and
      Hui, Kai  and
      Zhuang, Honglei  and
      Wu, Junru  and
      Yan, Le  and
      Shen, Jiaming  and
      Liu, Tianqi  and
      Liu, Jialu  and
      Metzler, Donald  and
      Wang, Xuanhui  and
      Bendersky, Michael",
    editor = "Duh, Kevin  and
      Gomez, Helena  and
      Bethard, Steven",
    booktitle = "Findings of the Association for Computational Linguistics: NAACL 2024",
    month = jun,
    year = "2024",
    address = "Mexico City, Mexico",
    publisher = "Association for Computational Linguistics",
    url = "https://aclanthology.org/2024.findings-naacl.97/",
    doi = "10.18653/v1/2024.findings-naacl.97",
    pages = "1504--1518"
}

@article{zhuang2025rank,
  title={Rank-R1: Enhancing Reasoning in LLM-based Document Rerankers via Reinforcement Learning},
  author={Zhuang, Shengyao and Ma, Xueguang and Koopman, Bevan and Lin, Jimmy and Zuccon, Guido},
  journal={arXiv preprint arXiv:2503.06034},
  year={2025}
}

@inproceedings{sun-etal-2023-chatgpt,
    title = "Is {C}hat{GPT} Good at Search? Investigating Large Language Models as Re-Ranking Agents",
    author = "Sun, Weiwei  and
      Yan, Lingyong  and
      Ma, Xinyu  and
      Wang, Shuaiqiang  and
      Ren, Pengjie  and
      Chen, Zhumin  and
      Yin, Dawei  and
      Ren, Zhaochun",
    editor = "Bouamor, Houda  and
      Pino, Juan  and
      Bali, Kalika",
    booktitle = "Proceedings of the 2023 Conference on Empirical Methods in Natural Language Processing",
    month = dec,
    year = "2023",
    address = "Singapore",
    publisher = "Association for Computational Linguistics",
    url = "https://aclanthology.org/2023.emnlp-main.923/",
    doi = "10.18653/v1/2023.emnlp-main.923",
    pages = "14918--14937"
}

@misc{gatto2025followupquestiongenerationenhanced,
      title={Follow-up Question Generation For Enhanced Patient-Provider Conversations}, 
      author={Joseph Gatto and Parker Seegmiller and Timothy Burdick and Inas S. Khayal and Sarah DeLozier and Sarah M. Preum},
      year={2025},
      eprint={2503.17509},
      archivePrefix={arXiv},
      primaryClass={cs.CL},
      url={https://arxiv.org/abs/2503.17509}, 
}

@InProceedings{pmlr-v126-si20a,
  title = 	 {Students Need More Attention: BERT-based Attention Model for Small Data with Application to Automatic Patient Message Triage},
  author =       {Si, Shijing and Wang, Rui and Wosik, Jedrek and Zhang, Hao and Dov, David and Wang, Guoyin and Carin, Lawrence},
  booktitle = 	 {Proceedings of the 5th Machine Learning for Healthcare Conference},
  pages = 	 {436--456},
  year = 	 {2020},
  editor = 	 {Doshi-Velez, Finale and Fackler, Jim and Jung, Ken and Kale, David and Ranganath, Rajesh and Wallace, Byron and Wiens, Jenna},
  volume = 	 {126},
  series = 	 {Proceedings of Machine Learning Research},
  month = 	 {07--08 Aug},
  publisher =    {PMLR},
  url = 	 {https://proceedings.mlr.press/v126/si20a.html}
}

@article{harzand2023clinician,
  title={Clinician-Trained Artificial Intelligence for Enhanced Routing of Patient Portal Messages in the Electronic Health Record},
  author={Harzand, Arash and Zia ul Haq, Muhammad and Hornback, Andrew M and Cowan, Alison D and Anderson, Blake},
  journal={medRxiv},
  pages={2023--11},
  year={2023},
  publisher={Cold Spring Harbor Laboratory Press}
}

@misc{mermin2023use,
  title={Use of Natural Language Processing of Patient-Initiated Electronic Health Record Messages to Identify Patients With COVID-19 Infection. JAMA Network Open 6, 7 (07 2023), e2322299--e2322299},
  author={Mermin-Bunnell, Kellen and Zhu, Yuanda and Hornback, Andrew and Damhorst, Gregory and Walker, Tiffany and others},
  year={2023}
}

@article{gatto2022identifying,
  title={Identifying the perceived severity of patient-generated telemedical queries regarding covid: Developing and evaluating a transfer learning--based solution},
  author={Gatto, Joseph and Seegmiller, Parker and Johnston, Garrett and Preum, Sarah Masud},
  journal={JMIR Medical Informatics},
  volume={10},
  number={9},
  pages={e37770},
  year={2022},
  publisher={JMIR Publications Toronto, Canada}
}

@article{nov2023putting,
  title={Putting ChatGPT’s medical advice to the (Turing) test: survey study},
  author={Nov, Oded and Singh, Nina and Mann, Devin},
  journal={JMIR Medical Education},
  volume={9},
  pages={e46939},
  year={2023},
  publisher={JMIR Publications Toronto, Canada}
}

@article{athavale2023potential,
  title={The potential of chatbots in chronic venous disease patient management},
  author={Athavale, Anand and Baier, Jonathan and Ross, Elsie and Fukaya, Eri},
  journal={JVS-vascular insights},
  volume={1},
  pages={100019},
  year={2023},
  publisher={Elsevier}
}

@inproceedings{lu-etal-2024-triageagent,
    title = "{T}riage{A}gent: Towards Better Multi-Agents Collaborations for Large Language Model-Based Clinical Triage",
    author = "Lu, Meng  and
      Ho, Brandon  and
      Ren, Dennis  and
      Wang, Xuan",
    editor = "Al-Onaizan, Yaser  and
      Bansal, Mohit  and
      Chen, Yun-Nung",
    booktitle = "Findings of the Association for Computational Linguistics: EMNLP 2024",
    month = nov,
    year = "2024",
    address = "Miami, Florida, USA",
    publisher = "Association for Computational Linguistics",
    url = "https://aclanthology.org/2024.findings-emnlp.329/",
    doi = "10.18653/v1/2024.findings-emnlp.329",
    pages = "5747--5764"
}

@article{liu2024survey,
  title={A survey on medical large language models: Technology, application, trustworthiness, and future directions},
  author={Liu, Lei and Yang, Xiaoyan and Lei, Junchi and Shen, Yue and Wang, Jian and Wei, Peng and Chu, Zhixuan and Qin, Zhan and Ren, Kui},
  journal={arXiv preprint arXiv:2406.03712},
  year={2024}
}

@article{tu2025towards,
  title={Towards conversational diagnostic artificial intelligence},
  author={Tu, Tao and Schaekermann, Mike and Palepu, Anil and Saab, Khaled and Freyberg, Jan and Tanno, Ryutaro and Wang, Amy and Li, Brenna and Amin, Mohamed and Cheng, Yong and others},
  journal={Nature},
  pages={1--9},
  year={2025},
  publisher={Nature Publishing Group UK London}
}

@inproceedings{gatto-etal-2025-follow,
    title = "Follow-up Question Generation For Enhanced Patient-Provider Conversations",
    author = "Gatto, Joseph  and
      Seegmiller, Parker  and
      Burdick, Timothy E.  and
      Khayal, Inas S.  and
      DeLozier, Sarah  and
      Preum, Sarah M.",
    editor = "Che, Wanxiang  and
      Nabende, Joyce  and
      Shutova, Ekaterina  and
      Pilehvar, Mohammad Taher",
    booktitle = "Proceedings of the 63rd Annual Meeting of the Association for Computational Linguistics (Volume 1: Long Papers)",
    month = jul,
    year = "2025",
    address = "Vienna, Austria",
    publisher = "Association for Computational Linguistics",
    url = "https://aclanthology.org/2025.acl-long.1226/",
    doi = "10.18653/v1/2025.acl-long.1226",
    pages = "25222--25240",
    ISBN = "979-8-89176-251-0"
}

@article{artsi2025large,
  title={Large language models in real-world clinical workflows: a systematic review of applications and implementation},
  author={Artsi, Yaara and Sorin, Vera and Glicksberg, Benjamin S and Korfiatis, Panagiotis and Nadkarni, Girish N and Klang, Eyal},
  journal={Frontiers in Digital Health},
  volume={7},
  pages={1659134},
  year={2025},
  publisher={Frontiers}
}

@article{chen2024huatuogpt,
  title={Huatuogpt-o1, towards medical complex reasoning with llms},
  author={Chen, Junying and Cai, Zhenyang and Ji, Ke and Wang, Xidong and Liu, Wanlong and Wang, Rongsheng and Hou, Jianye and Wang, Benyou},
  journal={arXiv preprint arXiv:2412.18925},
  year={2024}
}

@inproceedings{wang-etal-2025-towards-adapting,
    title = "Towards Adapting Open-Source Large Language Models for Expert-Level Clinical Note Generation",
    author = "Wang, Hanyin  and
      Gao, Chufan  and
      Liu, Bolun  and
      Xu, Qiping  and
      Hussein, Guleid  and
      Labban, Mohamad El  and
      Iheasirim, Kingsley  and
      Korsapati, Hariprasad Reddy  and
      Outcalt, Chuck  and
      Sun, Jimeng",
    editor = "Che, Wanxiang  and
      Nabende, Joyce  and
      Shutova, Ekaterina  and
      Pilehvar, Mohammad Taher",
    booktitle = "Findings of the Association for Computational Linguistics: ACL 2025",
    month = jul,
    year = "2025",
    address = "Vienna, Austria",
    publisher = "Association for Computational Linguistics",
    url = "https://aclanthology.org/2025.findings-acl.626/",
    doi = "10.18653/v1/2025.findings-acl.626",
    pages = "12084--12117",
    ISBN = "979-8-89176-256-5"
}

@article{naved2024contrasting,
  title={Contrasting rule and machine learning based digital self triage systems in the USA},
  author={Naved, Bilal A and Luo, Yuan},
  journal={NPJ digital medicine},
  volume={7},
  number={1},
  pages={381},
  year={2024},
  publisher={Nature Publishing Group UK London}
}

@article{apathy2024inbox,
  title={Inbox message prioritization and management approaches in primary care},
  author={Apathy, Nate C and Hicks, Katelyn and Bocknek, Lucy and Zabala, Garrett and Adams, Katharine and Gomes, Kylie M and Saggar, Tara},
  journal={JAMIA open},
  volume={7},
  number={4},
  pages={ooae135},
  year={2024},
  publisher={Oxford University Press}
}

@article{stillman2023death,
  title={Death by patient portal},
  author={Stillman, Michael},
  journal={JAMA},
  volume={330},
  number={3},
  pages={223--224},
  year={2023},
  publisher={American Medical Association}
}

@article{liu2025detecting,
  title={Detecting emergencies in patient portal messages using large language models and knowledge graph-based retrieval-augmented generation},
  author={Liu, Siru and Wright, Aileen P and McCoy, Allison B and Huang, Sean S and Steitz, Bryan and Wright, Adam},
  journal={Journal of the American Medical Informatics Association},
  volume={32},
  number={6},
  pages={1032--1039},
  year={2025},
  publisher={Oxford Academic}
}

@article{holmgren2025trends,
  title={Trends in physician electronic health record time and message volume},
  author={Holmgren, A Jay and Apathy, Nate C and Sinsky, Christine A and Adler-Milstein, Julia and Bates, David W and Rotenstein, Lisa},
  journal={JAMA Internal Medicine},
  volume={185},
  number={4},
  pages={461--463},
  year={2025},
  publisher={American Medical Association}
}

@article{apaydin2025secure,
  title={Secure Messages, Video Visits, and Burnout Among Primary Care Providers in the Veterans Health Administration: National Survey Study},
  author={Apaydin, Eric A and Der-Martirosian, Claudia and Yoo, Caroline and Rose, Danielle E and Jackson, Nicholas J and Stockdale, Susan E and Leung, Lucinda B},
  journal={Journal of Medical Internet Research},
  volume={27},
  pages={e68858},
  year={2025},
  publisher={JMIR Publications Toronto, Canada}
}

@misc{wornow2024zeroshotclinicaltrialpatient,
      title={Zero-Shot Clinical Trial Patient Matching with LLMs}, 
      author={Michael Wornow and Alejandro Lozano and Dev Dash and Jenelle Jindal and Kenneth W. Mahaffey and Nigam H. Shah},
      year={2024},
      eprint={2402.05125},
      archivePrefix={arXiv},
      primaryClass={cs.CL},
      url={https://arxiv.org/abs/2402.05125}, 
}

@inproceedings{brake-schaaf-2024-comparing,
    title = "Comparing Two Model Designs for Clinical Note Generation; Is an {LLM} a Useful Evaluator of Consistency?",
    author = "Brake, Nathan  and
      Schaaf, Thomas",
    editor = "Duh, Kevin  and
      Gomez, Helena  and
      Bethard, Steven",
    booktitle = "Findings of the Association for Computational Linguistics: NAACL 2024",
    month = jun,
    year = "2024",
    address = "Mexico City, Mexico",
    publisher = "Association for Computational Linguistics",
    url = "https://aclanthology.org/2024.findings-naacl.25/",
    doi = "10.18653/v1/2024.findings-naacl.25",
    pages = "352--363"
}

@inproceedings{zhuang2024setwise,
  title={A setwise approach for effective and highly efficient zero-shot ranking with large language models},
  author={Zhuang, Shengyao and Zhuang, Honglei and Koopman, Bevan and Zuccon, Guido},
  booktitle={Proceedings of the 47th International ACM SIGIR Conference on Research and Development in Information Retrieval},
  pages={38--47},
  year={2024}
}

@misc{xu2025surveymodelarchitecturesinformation,
      title={A Survey of Model Architectures in Information Retrieval}, 
      author={Zhichao Xu and Fengran Mo and Zhiqi Huang and Crystina Zhang and Puxuan Yu and Bei Wang and Jimmy Lin and Vivek Srikumar},
      year={2025},
      eprint={2502.14822},
      archivePrefix={arXiv},
      primaryClass={cs.IR},
      url={https://arxiv.org/abs/2502.14822}, 
}

@article{tanabe2004emergency,
  title={The Emergency Severity Index (version 3) 5-level triage system scores predict ED resource consumption},
  author={Tanabe, Paula and Gimbel, Rick and Yarnold, Paul R and Adams, James G},
  journal={Journal of Emergency Nursing},
  volume={30},
  number={1},
  pages={22--29},
  year={2004},
  publisher={Elsevier}
}

@inproceedings{lambert2025rewardbench,
  title={Rewardbench: Evaluating reward models for language modeling},
  author={Lambert, Nathan and Pyatkin, Valentina and Morrison, Jacob and Miranda, Lester James Validad and Lin, Bill Yuchen and Chandu, Khyathi and Dziri, Nouha and Kumar, Sachin and Zick, Tom and Choi, Yejin and others},
  booktitle={Findings of the Association for Computational Linguistics: NAACL 2025},
  pages={1755--1797},
  year={2025}
}

@article{bradley1952rank,
  title={Rank analysis of incomplete block designs: I. the method of paired comparisons},
  author={Bradley, Ralph Allan and Terry, Milton E},
  journal={Biometrika},
  volume={39},
  number={3/4},
  pages={324--345},
  year={1952},
  publisher={JSTOR}
}

@article{liu2025skywork,
  title={Skywork-Reward-V2: Scaling Preference Data Curation via Human-AI Synergy},
  author = {Liu, Chris Yuhao and Zeng, Liang and Xiao, Yuzhen and He, Jujie and Liu, Jiacai and Wang, Chaojie and Yan, Rui and Shen, Wei and Zhang, Fuxiang and Xu, Jiacheng and Liu, Yang and Zhou, Yahui},
  journal={arXiv preprint arXiv:2507.01352},
  year={2025}
}

@misc{qwen3technicalreport,
      title={Qwen3 Technical Report}, 
      author={Qwen Team},
      year={2025},
      eprint={2505.09388},
      archivePrefix={arXiv},
      primaryClass={cs.CL},
      url={https://arxiv.org/abs/2505.09388}, 
}

@misc{sellergren2025medgemmatechnicalreport,
      title={MedGemma Technical Report}, 
      author={Andrew Sellergren and Sahar Kazemzadeh and Tiam Jaroensri and Atilla Kiraly and Madeleine Traverse and Timo Kohlberger and Shawn Xu and Fayaz Jamil and Cían Hughes and Charles Lau and Justin Chen and Fereshteh Mahvar and Liron Yatziv and Tiffany Chen and Bram Sterling and Stefanie Anna Baby and Susanna Maria Baby and Jeremy Lai and Samuel Schmidgall and Lu Yang and Kejia Chen and Per Bjornsson and Shashir Reddy and Ryan Brush and Kenneth Philbrick and Mercy Asiedu and Ines Mezerreg and Howard Hu and Howard Yang and Richa Tiwari and Sunny Jansen and Preeti Singh and Yun Liu and Shekoofeh Azizi and Aishwarya Kamath and Johan Ferret and Shreya Pathak and Nino Vieillard and Ramona Merhej and Sarah Perrin and Tatiana Matejovicova and Alexandre Ramé and Morgane Riviere and Louis Rouillard and Thomas Mesnard and Geoffrey Cideron and Jean-bastien Grill and Sabela Ramos and Edouard Yvinec and Michelle Casbon and Elena Buchatskaya and Jean-Baptiste Alayrac and Dmitry Lepikhin and Vlad Feinberg and Sebastian Borgeaud and Alek Andreev and Cassidy Hardin and Robert Dadashi and Léonard Hussenot and Armand Joulin and Olivier Bachem and Yossi Matias and Katherine Chou and Avinatan Hassidim and Kavi Goel and Clement Farabet and Joelle Barral and Tris Warkentin and Jonathon Shlens and David Fleet and Victor Cotruta and Omar Sanseviero and Gus Martins and Phoebe Kirk and Anand Rao and Shravya Shetty and David F. Steiner and Can Kirmizibayrak and Rory Pilgrim and Daniel Golden and Lin Yang},
      year={2025},
      eprint={2507.05201},
      archivePrefix={arXiv},
      primaryClass={cs.AI},
      url={https://arxiv.org/abs/2507.05201}, 
}

@misc{openai2025gptoss120bgptoss20bmodel,
      title={gpt-oss-120b \& gpt-oss-20b Model Card}, 
      author={OpenAI and : and Sandhini Agarwal and Lama Ahmad and Jason Ai and Sam Altman and Andy Applebaum and Edwin Arbus and Rahul K. Arora and Yu Bai and Bowen Baker and Haiming Bao and Boaz Barak and Ally Bennett and Tyler Bertao and Nivedita Brett and Eugene Brevdo and Greg Brockman and Sebastien Bubeck and Che Chang and Kai Chen and Mark Chen and Enoch Cheung and Aidan Clark and Dan Cook and Marat Dukhan and Casey Dvorak and Kevin Fives and Vlad Fomenko and Timur Garipov and Kristian Georgiev and Mia Glaese and Tarun Gogineni and Adam Goucher and Lukas Gross and Katia Gil Guzman and John Hallman and Jackie Hehir and Johannes Heidecke and Alec Helyar and Haitang Hu and Romain Huet and Jacob Huh and Saachi Jain and Zach Johnson and Chris Koch and Irina Kofman and Dominik Kundel and Jason Kwon and Volodymyr Kyrylov and Elaine Ya Le and Guillaume Leclerc and James Park Lennon and Scott Lessans and Mario Lezcano-Casado and Yuanzhi Li and Zhuohan Li and Ji Lin and Jordan Liss and Lily and Liu and Jiancheng Liu and Kevin Lu and Chris Lu and Zoran Martinovic and Lindsay McCallum and Josh McGrath and Scott McKinney and Aidan McLaughlin and Song Mei and Steve Mostovoy and Tong Mu and Gideon Myles and Alexander Neitz and Alex Nichol and Jakub Pachocki and Alex Paino and Dana Palmie and Ashley Pantuliano and Giambattista Parascandolo and Jongsoo Park and Leher Pathak and Carolina Paz and Ludovic Peran and Dmitry Pimenov and Michelle Pokrass and Elizabeth Proehl and Huida Qiu and Gaby Raila and Filippo Raso and Hongyu Ren and Kimmy Richardson and David Robinson and Bob Rotsted and Hadi Salman and Suvansh Sanjeev and Max Schwarzer and D. Sculley and Harshit Sikchi and Kendal Simon and Karan Singhal and Yang Song and Dane Stuckey and Zhiqing Sun and Philippe Tillet and Sam Toizer and Foivos Tsimpourlas and Nikhil Vyas and Eric Wallace and Xin Wang and Miles Wang and Olivia Watkins and Kevin Weil and Amy Wendling and Kevin Whinnery and Cedric Whitney and Hannah Wong and Lin Yang and Yu Yang and Michihiro Yasunaga and Kristen Ying and Wojciech Zaremba and Wenting Zhan and Cyril Zhang and Brian Zhang and Eddie Zhang and Shengjia Zhao},
      year={2025},
      eprint={2508.10925},
      archivePrefix={arXiv},
      primaryClass={cs.CL},
      url={https://arxiv.org/abs/2508.10925}, 
}

@misc{hu2021loralowrankadaptationlarge,
      title={LoRA: Low-Rank Adaptation of Large Language Models}, 
      author={Edward J. Hu and Yelong Shen and Phillip Wallis and Zeyuan Allen-Zhu and Yuanzhi Li and Shean Wang and Lu Wang and Weizhu Chen},
      year={2021},
      eprint={2106.09685},
      archivePrefix={arXiv},
      primaryClass={cs.CL},
      url={https://arxiv.org/abs/2106.09685}, 
}

@inproceedings{ranx,
  author       = {Elias Bassani},
  title        = {ranx: {A} Blazing-Fast Python Library for Ranking Evaluation and Comparison},
  booktitle    = {{ECIR} {(2)}},
  series       = {Lecture Notes in Computer Science},
  volume       = {13186},
  pages        = {259--264},
  publisher    = {Springer},
  year         = {2022},
  doi          = {10.1007/978-3-030-99739-7\_30}
}

@article{DBLP:journals/tois/JarvelinK02,
    author    = {Kalervo J{\"{a}}rvelin and
                Jaana Kek{\"{a}}l{\"{a}}inen},
    title     = {Cumulated gain-based evaluation of {IR} techniques},
    journal   = {{ACM} Trans. Inf. Syst.},
    volume    = {20},
    number    = {4},
    pages     = {422--446},
    year      = {2002}
}

@article{10.1093/jamiaopen/ooae078,
    author = {Yang, Jie and So, Jonathan and Zhang, Hao and Jones, Simon and Connolly, Denise M and Golding, Claudia and Griffes, Esmelin and Szerencsy, Adam C and Wu, Tzer (Jason) and Aphinyanaphongs, Yindalon and Major, Vincent J},
    title = {Development and evaluation of an artificial intelligence-based workflow for the prioritization of patient portal messages},
    journal = {JAMIA Open},
    volume = {7},
    number = {3},
    pages = {ooae078},
    year = {2024},
    month = {08},
    issn = {2574-2531},
    doi = {10.1093/jamiaopen/ooae078},
    url = {https://doi.org/10.1093/jamiaopen/ooae078},
    eprint = {https://academic.oup.com/jamiaopen/article-pdf/7/3/ooae078/58834078/ooae078_supplementary_data.pdf},
}

@inproceedings{devlin-etal-2019-bert,
    title = "{BERT}: Pre-training of Deep Bidirectional Transformers for Language Understanding",
    author = "Devlin, Jacob  and
      Chang, Ming-Wei  and
      Lee, Kenton  and
      Toutanova, Kristina",
    editor = "Burstein, Jill  and
      Doran, Christy  and
      Solorio, Thamar",
    booktitle = "Proceedings of the 2019 Conference of the North {A}merican Chapter of the Association for Computational Linguistics: Human Language Technologies, Volume 1 (Long and Short Papers)",
    month = jun,
    year = "2019",
    address = "Minneapolis, Minnesota",
    publisher = "Association for Computational Linguistics",
    url = "https://aclanthology.org/N19-1423/",
    doi = "10.18653/v1/N19-1423",
    pages = "4171--4186"
}

@inproceedings{ozkaynak2014examining,
  title={Examining the multi-level fit between work and technology in a secure messaging implementation},
  author={Ozkaynak, Mustafa and Johnson, Sharon and Shimada, Stephanie and Petrakis, Beth Ann and Tulu, Bengisu and Archambeault, Cliona and Fix, Gemmae and Schwartz, Erin and Woods, Susan},
  booktitle={AMIA Annual Symposium Proceedings},
  volume={2014},
  pages={954},
  year={2014}
}

@article{shah2018simple,
  title={Simple, robust and optimal ranking from pairwise comparisons},
  author={Shah, Nihar B and Wainwright, Martin J},
  journal={Journal of machine learning research},
  volume={18},
  number={199},
  pages={1--38},
  year={2018}
}

@misc{crowdcent_symmetric_ndcg_2025,
  author       = {{CrowdCent}},
  title        = {Scoring: Symmetric NDCG@k},
  year         = {2025},
  month        = dec,
  day          = {30},
  url          = {https://docs.crowdcent.com/scoring/},
  note         = {CrowdCent Challenge Docs. Accessed: 2026-01-02}
}

@inproceedings{10.1145/3340531.3412123,
author = {Gienapp, Lukas and Fr\"{o}be, Maik and Hagen, Matthias and Potthast, Martin},
title = {The Impact of Negative Relevance Judgments on NDCG},
year = {2020},
isbn = {9781450368599},
publisher = {Association for Computing Machinery},
address = {New York, NY, USA},
url = {https://doi.org/10.1145/3340531.3412123},
doi = {10.1145/3340531.3412123},
booktitle = {Proceedings of the 29th ACM International Conference on Information \& Knowledge Management},
pages = {2037–2040},
numpages = {4},
location = {Virtual Event, Ireland},
series = {CIKM '20}
}

@article{adler2020electronic,
  title={Electronic health records and burnout: time spent on the electronic health record after hours and message volume associated with exhaustion but not with cynicism among primary care clinicians},
  author={Adler-Milstein, Julia and Zhao, Wendi and Willard-Grace, Rachel and Knox, Margae and Grumbach, Kevin},
  journal={Journal of the American Medical Informatics Association},
  volume={27},
  number={4},
  pages={531--538},
  year={2020},
  publisher={Oxford University Press}
}

@inproceedings{mcsherry2008computing,
  title={Computing information retrieval performance measures efficiently in the presence of tied scores},
  author={McSherry, Frank and Najork, Marc},
  booktitle={European conference on information retrieval},
  pages={414--421},
  year={2008},
  organization={Springer}
}

@misc{vonwerra2022trl,
  author = {Leandro von Werra and Younes Belkada and Lewis Tunstall and Edward Beeching and Tristan Thrush and Nathan Lambert and Shengyi Huang and Kashif Rasul and Quentin Gallouédec},
  title = {TRL: Transformer Reinforcement Learning},
  year = {2020},
  publisher = {GitHub},
  journal = {GitHub repository},
  howpublished = {\url{https://github.com/huggingface/trl}}
}

@inproceedings{qlora,
author = {Dettmers, Tim and Pagnoni, Artidoro and Holtzman, Ari and Zettlemoyer, Luke},
title = {QLORA: efficient finetuning of quantized LLMs},
year = {2023},
publisher = {Curran Associates Inc.},
address = {Red Hook, NY, USA},
booktitle = {Proceedings of the 37th International Conference on Neural Information Processing Systems},
articleno = {441},
numpages = {28},
location = {New Orleans, LA, USA},
series = {NIPS '23}
}

@misc{unsloth,
  author = {Daniel Han, Michael Han and Unsloth team},
  title = {Unsloth},
  url = {http://github.com/unslothai/unsloth},
  year = {2023}
}

@inproceedings{wolf-etal-2020-transformers,
    title = "Transformers: State-of-the-Art Natural Language Processing",
    author = "Thomas Wolf and Lysandre Debut and Victor Sanh and Julien Chaumond and Clement Delangue and Anthony Moi and Pierric Cistac and Tim Rault and Rémi Louf and Morgan Funtowicz and Joe Davison and Sam Shleifer and Patrick von Platen and Clara Ma and Yacine Jernite and Julien Plu and Canwen Xu and Teven Le Scao and Sylvain Gugger and Mariama Drame and Quentin Lhoest and Alexander M. Rush",
    booktitle = "Proceedings of the 2020 Conference on Empirical Methods in Natural Language Processing: System Demonstrations",
    month = oct,
    year = "2020",
    address = "Online",
    publisher = "Association for Computational Linguistics",
    url = "https://www.aclweb.org/anthology/2020.emnlp-demos.6",
    pages = "38--45"
}

@misc{liang2023holisticevaluationlanguagemodels,
      title={Holistic Evaluation of Language Models}, 
      author={Percy Liang and Rishi Bommasani and Tony Lee and Dimitris Tsipras and Dilara Soylu and Michihiro Yasunaga and Yian Zhang and Deepak Narayanan and Yuhuai Wu and Ananya Kumar and Benjamin Newman and Binhang Yuan and Bobby Yan and Ce Zhang and Christian Cosgrove and Christopher D. Manning and Christopher Ré and Diana Acosta-Navas and Drew A. Hudson and Eric Zelikman and Esin Durmus and Faisal Ladhak and Frieda Rong and Hongyu Ren and Huaxiu Yao and Jue Wang and Keshav Santhanam and Laurel Orr and Lucia Zheng and Mert Yuksekgonul and Mirac Suzgun and Nathan Kim and Neel Guha and Niladri Chatterji and Omar Khattab and Peter Henderson and Qian Huang and Ryan Chi and Sang Michael Xie and Shibani Santurkar and Surya Ganguli and Tatsunori Hashimoto and Thomas Icard and Tianyi Zhang and Vishrav Chaudhary and William Wang and Xuechen Li and Yifan Mai and Yuhui Zhang and Yuta Koreeda},
      year={2023},
      eprint={2211.09110},
      archivePrefix={arXiv},
      primaryClass={cs.CL},
      url={https://arxiv.org/abs/2211.09110}, 
}

\newpage
\appendix
\section{Additional Data Details}

\subsection{Inter-Annotator Agreement}

In Table \ref{tab:iaa_metrics} we show the inter-annotator agreement metrics for PMR-Synth. 
\begin{table}[!h]
\centering
\small 
\begin{tabular}{lcc}
\hline
\textbf{Metric} & \textbf{Value} & \textbf{N (item pairs)} \\
\hline
Percent agreement & 0.85 & 20 \\
Krippendorff's $\alpha$ & 0.63 & 20 \\
\hline
\end{tabular}
\caption{Inter-annotator agreement metrics (two annotators, binary "Which patient is more urgent" labels). Across all 20 samples, there were only three disagreements between the two annotators, showing strong alignment and annotation quality. }\label{tab:iaa_metrics}
\end{table}

\subsection{Extrinsic Eval Data Distribution}\label{sec:extrinsic_eval}

\begin{table}[h]
\centering
\small
\begin{tabular}{c|ccc}
\textbf{Level} & \textbf{Reddit} & \textbf{Synth} & \textbf{Real} \\ \hline
1 & 5 & 5 & 5 \\
2 & 5 & 5 & 3 \\
3 & 5 & 5 & 5 \\
4 & 5 & 5 & 7 \\
5 & 5 & 5 & 7 \\
6 & 5 & 5 & 4 \\
\end{tabular}
\caption{Extrinsic Evaluation Inbox Distribution by Level. }\label{tab:extrinsic_dist}
\end{table}

In Table \ref{tab:extrinsic_dist} we show the number of samples per-level in the extrinsic inbox evaluations. We note that to ensure data quality in PMR-Real, the authors manually selected each sample included in the inbox, verifying that each extracted label closely aligns with the label definition. This was performed as (i) our secure computing server only has access to smaller LLMs, making automated label extraction more challenging and (ii) unlike pairwise comparisons, we are not able to do LLM post-processing where we confirm, using the clinicians response, that one sample is more urgent than another. In general, we tried to keep the label distribution even, but given the manual efforts it was challenging to maintain the uniformity without extending manual data review beyond capacity. 

\subsection{Dataset Examples}
In Table \ref{tab:label_definitions} we show each class label, it's definition, and an example to help better understand our dataset.

\begin{table*}[ht]
    \centering
    \small
    \begin{tabular}{
        >{\raggedright\arraybackslash}p{1cm}
        >{\raggedright\arraybackslash}p{4cm}
        >{\raggedright\arraybackslash}p{4cm}
        >{\raggedright\arraybackslash}p{4cm}
    }
        \toprule
        Level & Definition Summary & Example Message & Example Response \\
        \midrule
        1 &  Patient is facing life-threatening symptoms and must seek emergency care. & Hey doc, having really bad shortness of breath today & You should be seen at the ED immediately!  \\
        \midrule
        2 &  Patient is experiencing urgent yet non-life-threatening symptoms worthy of same-day care, such as urgent care or a walk-in clinic. & Hi doc, have a fever of 102 and some spots in my throat.  &  You may have strep and should be seen at an urgent care ASAP. \\
        \midrule
        3 & Patient is experiencing semi-urgent symptoms worthy of an in-person healthcare visit. The visit must occur in the next 1-3 days. & Hi Doc, I fell down a few days ago and have been experiencing this lingering back pain. & I can see you tomorrow if you are available. Thanks. \\
        \midrule
        4 & Patient is experiencing non-urgent symptoms worthy of an in-person healthcare visit. The visit is not time-sensitive. & Been struggling with this cough off-and-on for a month now. Wondering if we can meet to chat. & Sure, we have opening for next week. Please call the office. \\
        \midrule
        5 & Patient is experiencing symptoms that may be alleviated by self-treatment options. They do not require an in-person visit but do require remote guidance. & Hi doc, got a stuffy nose, thought I would let you know. Any meds you suggest? & You can take sudafed for the congestion. Let me know if it progresses. \\
        \midrule
        6 & Patient is experiencing symptoms that require no medical attention and are not of concern. No further action is needed.  & Wondering where I can get a COVID test these days? Been feeling a little sniffly. & You can get a COVID test from a local drugstore. \\
        \bottomrule
    \end{tabular}
    \caption{Label definitions with examples. Due to space constraints, we use manually-written examples to communicate the label space. In practice, messages and responses are much more complex. }
    \label{tab:label_definitions}
\end{table*}

\subsection{Urgency Annotation Details} \label{apn:labeling_details}

In this section, we detail data filtration steps taken to curate high-quality patient messages for PMR-Real and PMR-Reddit. Note that these steps do not apply to PMR-Synth as those messages are hand-crafted and not sourced from a larger corpus. 

\subsubsection{Clinician's Response Classification} 

When classifying clinicians' responses to messages into our 6-level ordinal scale, we allow the model to predict two additional classes which we use for data filtering. The first is ``UNCLEAR", which the model can select if the response does not cleanly fit into one of our categories. The next is ``SUPPORTIVE CARE" which captures suggestions for things like  physical therapy or non-urgent mental health sessions. We found that filtering these two sample types was necessary for data quality as (i) any sample which is unclear should not be included and (ii) suggestions to seek supportive care are often of a  fundamentally different nature compared to suggestions to seek to acute care. Without this label, the model may conflate suggestions to visit one's PCP and one's physical therapist, for example, as the same urgency tier, which is not correct given our task definition.

\subsubsection{Data Quality Checks}\label{apn:data_quality}

For a given data source with paired (message, clinician's response) data tuples, we can automatically extract an urgency label from the clinician's response based on urgency hierarchy shown in Table \ref{tab:label_definitions}. We then create pairwise classification pairs by sampling messages with different urgency labels. For example, we may sample a Level 1 message and a Level 3 messages. Since level 1 is the highest urgency, we can create a pair where we know the correct label for ``which message is more medically urgent". 

To ensure the quality of the pairwise annotations, we perform additional data quality checks before creating our test sets. For PMR-Reddit and PMR-Real, we pass each sampled data pair through an LLM --- providing both the message and clinician's response for each sample. The prompt asks the model to review which of the two patients is more urgent based on the clinician's response. Importantly, the model can choose one of the patients \textbf{or} decide that it is unclear which patient is more urgent. We only retain pairs for which the LLM (i) agrees with our auto-label and (ii) does not find the decision to be unclear. We perform this filtration twice with two different prompt variations, only keeping pairs where both filters find it clear from the clinician's response which of the two patients is more urgent.  

\subsubsection{Sample Inclusion Criteria}

\paragraph{PMR-Reddit: } Using LLMs as filtration mechanisms, we run a series of filters over the raw PMR-Reddit samples to identify patient messages that meet desirable criteria. Specifically, we ensure that each patient is over 18 years old and that they are suffering from an acutely onset issue. 

\paragraph{PMR-Real: } We similarly filter for patients who are 18 years or older, but without use of an LLM as we can leverage structured EHR data. Then, we filter for patients suffering acutely onset issues.

\subsection{Additional PMR-Synth Details}\label{sec:pmr_synth}

\subsubsection{Annotation}
We constructed two separate inboxes of synthetic patient portal messages of size n=30 to create PMR-Synth. We can build all possible $n\choose2$ pairs of messages and gave them to a team of triage nurses and physicians at our collaborating hospital for annotation. Annotators were instructed to select which of two patients were more medically urgent based on the message and structured EHR data presented to them. If the annotators found any samples to be challenging, a 2nd annotator could be requested for additional review. Due to the significant annotation expense, we were unable to do a full-scale 2-annotations per pair. However, we did find strong inter-annotator agreement (see Table \ref{tab:iaa_metrics}) on a subset of the samples between two expert annotators. 

Each annotator was recruited through our collaboration with a partner hospital and was compensated \$50-per-hour for their annotation efforts. Each annotator was aware of the data's intended use. Each annotator was verbally instructed on how to perform the annotation task, which as mentioned above entailed choosing which of two patients they would provide priority medical care to. They were further instructed not to assign any ties and to use the corresponding EHR data in whichever way they see fit. From our discussions with annotators, we can note that in the case where the degree of medical urgency between two patients was very similar, the EHR data often served as a useful tie-breaking mechanism to help decide which patient was generally more high-risk. 

\subsubsection{Inbox Creation}

Our extrinsic evaluation metric requires relevancy labels which for PMR-Reddit and PMR-Real were extracted from ground-truth responses. While we do not have ground-truth responses for PMR-Synth, we do have the ability to sort the 30 message inbox by win-rate and assign a discrete label to each element in the inbox based on the position it was sorted to (e.g. Top 1/6 most urgent message receive Level 1, bottom 1/6 receive Level 6). We use this strategy to align per-difficulty PMR-Synth results with PMR-Reddit and PMR-Real.

\section{Additional Training Details}\label{apn:training_deets}

All fine-tuned models are trained using LoRA/QLoRA \cite{hu2021loralowrankadaptationlarge, qlora} with rank = 64 and alpha = 64. For all Qwen and MedGemma baselines we use Unsloth \cite{unsloth} 4-bit BitsAndBytes \footnote{https://github.com/bitsandbytes-foundation/bitsandbytes} quantizations. As the SkyWork \cite{liu2025skywork} reward models could fit on our secure computing platform we used the original  model for such experiments. We use TRL \cite{vonwerra2022trl} and Huggingface Transformeres \cite{wolf-etal-2020-transformers} for model development. 

All PMR-Reddit and Real models are trained for 1 epoch with learning rate 1e-5 with the best model chosen from the development set. For PMR-Synth, since our annotation was for all $n\choose2$ samples in an inbox, we cannot create a development set with disjoint patient messages. Due to this dataset having far fewer unique queries, we ran two experiments, one with one epoch and another with three epochs. We found training longer to be more effective and reported the 3 epoch result for all trained models with no other parameter search. 

In Table \ref{tab:training_data} we show the  distribution of training samples for PMR-Reddit and PMR-Synth. Specifically, we take the following steps when creating SFT and Reward training samples: 

\begin{enumerate}
    \item We random sample an anchor sample. This is any sample with an urgency level between 2-5 (as Level 1 and Level 6 samples cannot have anything reliably considered to be ``more" or ``less" urgent respectively).
    \item We then sample two messages, one which is more and one which is less urgent than the anchor sample. This creates a triplet of (anchor, more urgent, less urgent) messages. 
    \begin{enumerate}
        \item Note that we cap how many times a sample can be included as a more or less urgent training instance, as performing the full $n \choose 2$ would be very computationally expensive for our larger training sets (i.e. PMR-Reddit, PMR-Real). 
    \end{enumerate}
    \item Now that we have this triplet, we use it to build SFT and Reward training samples. 
    \begin{enumerate}
        \item \textbf{SFT: } We create four samples from each triplet. Recall that for SFT, we are prompting for a pair of patients (a,b), to determine if b is more urgent than a. Thus, we create the training samples (anchor, more urgent, yes), (anchor, less urgent, no), (more urgent, anchor, no), (less urgent, anchor, yes). 
        \item \textbf{Reward: } Recall that UrgentReward asks the model to write a sample more urgent than the one provided. Thus, we create a training triplet (anchor, more urgent, less urgent) which corresponds to a positive and negative completion to the prompt. Because of the disproportionate number of SFT samples, we also add the inverse prompt (i.e. write a message which \textit{less} urgent) to the training pool and provide the triplet (anchor, less urgent, more urgent). This design choice (i) helps balance out the number of gradient updates compared to SFT and (ii) provides the same advantage SFT receives in learning from the inverse. Note we never apply the inverse prompt during test time. 
        \end{enumerate}
\end{enumerate}

\begin{table}[!h]
\centering
\small 
\begin{tabular}{lrr}
\toprule
 & PMR-Reddit & PMR-Synth \\
\midrule
 Unique Message Count & 772 & 30 \\
\midrule
UrgentSFT & 26,544 & 2,352 \\
UrgentReward & 13,272 & 1,164 \\
\bottomrule
\end{tabular}
\caption{Amount of training pairs/triplets used for UrgentSFT and UrgentReward respectively. Training pairs are built from the set of unique messages for which a messages has been classified into our 6-level hierarchy. UrgentSFT has 2x the training data because for a given triplet (Anchor, More Urgent, Less Urgent) used to train UrgentReward, we convert this into (Anchor, More Urgent) and (Anchor, Less Urgent) SFT pairs. However, the data and number of comparisons between unique samples is the same. }
\label{tab:training_data}
\end{table}

\section{Additional Results}\label{sec:additional_results}

\subsection{Training Set Size Ablation}\label{sec:training_size_ablation}
\begin{table}[!h]
\small
\centering
\begin{tabular}{lllll}
\toprule
Model & Easy & Med & Hard & Total \\
\midrule
UrgentSFT-Q4B-Small & 0.82 & 0.66 & 0.61 & 0.68 \\
UrgentSFT-Q4B-Large & 0.84 & 0.69 & 0.63 & 0.70 \\
UrgentReward-4B-Small & 0.91 & 0.75 & 0.69 & 0.77 \\
UrgentReward-4B-Large & 0.91 & 0.80 & 0.80 & 0.82 \\
\midrule 
UrgentSFT-Q8B-Small & 0.84 & 0.68 & 0.59 & 0.69 \\
UrgentSFT-Q8B-Large & 0.90 & 0.75 & 0.69 & 0.76 \\
UrgentReward-8B-Small & 0.93 & 0.81 & 0.81 & 0.84 \\
UrgentReward-8B-Large & 0.93 & 0.82 & 0.85 & 0.85 \\
\midrule 
UrgentSFT-Q32B-Small & 0.96 & 0.82 & 0.77 & 0.84 \\
UrgentSFT-Q32B-Large & 0.96 & 0.86 & 0.84 & 0.87 \\
\midrule 
UrgentSFT-M27B-Small & 0.95 & 0.83 & 0.82 & 0.85 \\
UrgentSFT-M27B-Large & 0.98 & 0.85 & 0.87 & 0.88 \\
\bottomrule
\end{tabular}
\caption{Comparing the performance of each model on \textbf{PMR-Reddit} when presented with a smaller training set ($\approx 2,300$ triplets) vs a larger training set ($\approx 6,600$ triplets). In this table, \underline{Q} denotes a Qwen3 model and \underline{M} denotes a MedGemma model. We find that UrgentReward models are \textit{more sample efficient} and can make 4B and 8B parameter models viable. Note that the results in Table \ref{tab:pairwise-results} display the larger dataset. }
\label{tab:ablation_reddit_data_size}
\end{table}

In Table \ref{tab:ablation_reddit_data_size} we demonstrate how number of training samples affects pairwise classification performance on PMR-Reddit.

\subsection{EHR Ablation}\label{sec:ehr_ablation}
\begin{table}[htbp]
\centering
\small
\begin{tabular}{lllll}
\toprule
Model & Easy & Med & Hard & Total \\
\midrule
\textit{UrgentSFT-Qwen3-32B} &&&& \\
With EHR & 0.92 & 0.70 & 0.60 & 0.69 \\
No EHR   & 0.88 & 0.72 & 0.62 & 0.70 \\
\midrule
\textit{UrgentSFT-MedGemma} &&&& \\
With EHR & 0.93 & 0.77 & 0.60 & 0.73 \\
No EHR   & 0.92 & 0.74 & 0.57 & 0.70 \\
\midrule 
\textit{UrgentReward-8B} &&&& \\
With EHR & 0.91 & 0.73 & 0.62 & 0.71 \\
No EHR   & 0.93 & 0.77 & 0.57 & 0.71 \\
\midrule 
\textit{GPT-OSS-120B} &&&& \\
With EHR & 0.79 & 0.60 & 0.50 & 0.59 \\
No EHR   & 0.85 & 0.66 & 0.44 & 0.62 \\
\bottomrule
\end{tabular}
\caption{Showing model performance on PMR-Synth with and without inclusion of the structured EHR data as part of the input. }
\label{tab:with_and_without}
\end{table}

In Table \ref{tab:with_and_without} we show an ablation where we train with and without the EHR in the input for PMR-Synth. Interestingly, our Qwen-based models (i.e. UrgentSFT-Qwen32B and UrgentReward-8B) find little to no benefit from the inclusion of the EHR data. This likely suggests that the model is focusing its attention on the message, limiting it's capacity to achieve higher performance by using all available information. 

\begin{table}[!h]
    \centering
    \small
    \begin{tabular}{l|c}
        \toprule
        Category & Count (\%) \\
        \midrule
        Overlapping Correct (Both correct) & 239 (54.9\%) \\
        Overlapping Incorrect (Both incorrect) & 102 (23.4\%) \\
        OSS correct, MedGemma incorrect & 17 (3.9\%) \\
        OSS incorrect, MedGemma correct & 77 (17.7\%) \\
        \midrule
        \textbf{Total} & \textbf{435} \\
        \bottomrule
    \end{tabular}
    \caption{Overlap analysis of prediction outcomes between OSS and MedGemma-SFT models. The majority of predictions are either correctly classified by both models or incorrectly classified by both, while MedGemma-SFT corrects a larger fraction of errors made by OSS than vice versa.}
    \label{tab:prediction_overlap}
\end{table}

However, we do see a clear boost in performance from the UrgentSFT-MedGemma model. This is intuitive as such models should be more naturally pre-disposed to medical terminology, making the EHR data more useful. 

Finally, we notice that removing the EHR \textit{improves} performance for GPT-OSS --- specifically on easy and medium samples. Given that easy samples, for example, may be more likely to have their urgency label be independent of the EHR, it is not surprising that the model with no in-domain training found the structured EHR data distracting. Future works may wish to explore extending medical reasoning capacity to process structured EHR information before making a prediction. 

\subsection{Multi-Class Inference and Robustness to Initial Order}\label{apn:multi_class_instability}
\begin{table*}
\centering
\small
\begin{tabular}{lcc|cc}
\toprule
 & Avg T-NDCG @ 10 & SD @ 10 & Avg T-NDCG @ 30 & SD @ 30 \\
 \textbf{PMR-Reddit} &  &  &  &  \\
MedGemma-27B & 0.49 & 0.08 & 0.25 & 0.05 \\
MedGemma-27B-SFT & 0.62 & 0.06 & 0.32 & 0.01 \\
Qwen3-32B & 0.59 & 0.06 & 0.28 & 0.03 \\
Qwen3-32B-SFT & 0.50 & 0.08 & 0.24 & 0.04 \\
GPT-OSS & 0.40 & 0.09 & 0.18 & 0.05 \\
\midrule
\textbf{PMR-Synth} &  &  &  &  \\
MedGemma-27B & 0.54 & 0.08 & 0.24 & 0.06 \\
MedGemma-27B+SFT & 0.27 & 0.08 & 0.18 & 0.03 \\
Qwen3-32B & 0.5 & 0.08 & 0.26 & 0.03 \\
Qwen3-32B + SFT & 0.57 & 0.05 & 0.27 & 0.03 \\
GPT-OSS & 0.48 & 0.08 & 0.23 & 0.04 \\
\bottomrule
\end{tabular}
\caption{Average NDCG over 300 trials for the PMR-Synth MultiClass. These results correspond to those found in Table \ref{tab:pairwise-results}. }
\label{tab:exp_ndcg_2}
\end{table*}
The NDCG metric will not break ties, which means that a multi-class model's performance will somewhat depend on a meaningless order within each discrete class. This is a big issue for multiclass when the LLM decides to over-predict one of the classes. One way to address this problem is to shuffle the order of the elements within a discrete class multiple times and report the average metric. This is exactly what we do in Table \ref{tab:pairwise-results} of the main paper. Here in Table \ref{tab:exp_ndcg_2}  we show that while some multi-class models can achieve reasonable T-NDCG scores, the performance can vary greatly depending on the intra-class shuffling. Thus, our methods not only have stronger performance, but more consistent rankings.

\section{Error Analysis}\label{apn:error_analysis_bias}

\begin{table*}[!h]
\centering
\small
\begin{tabular}{lcccc}
\toprule
Model & Urgency Level & Gender & Incorrect & Correct \\
 &  &  &  &  \\
\midrule
OSS & More Urgent & Female & 150 (41.9\%) & 208 (58.1\%) \\
OSS & More Urgent & Male   & 29 (37.7\%)  & 48 (62.3\%) \\
OSS & Less Urgent & Female & 133 (36.2\%) & 234 (63.8\%) \\
OSS & Less Urgent & Male   & 46 (67.6\%)  & 22 (32.4\%) \\
MedGemma-SFT & More Urgent & Female & 101 (28.2\%) & 257 (71.8\%) \\
MedGemma-SFT & More Urgent & Male   & 18 (23.4\%)  & 59 (76.6\%) \\
MedGemma-SFT & Less Urgent & Female & 92 (25.1\%)  & 275 (74.9\%) \\
MedGemma-SFT & Less Urgent & Male   & 27 (39.7\%)  & 41 (60.3\%) \\
\bottomrule
\end{tabular}
\caption{Prediction performance stratified by urgency level and gender. In other words, when the more urgent patient was Male/Female, what was the performance? MedGemma-SFT consistently achieves higher accuracy than OSS across all urgency–gender combinations, with particularly strong improvements for female cases. For example, the row for "OSS, More Urgent, Female" shows that when the more urgent patient is female, OSS correctly classified 208 cases (58.1\%) and incorrectly classified 150 cases (41.9\%).}
\label{tab:urgency_gender}
\end{table*}

\begin{table*}[!h]
    \centering
    \small
    \begin{tabular}{l|lccc}
        \toprule
        Model & Urgency Level & Correct & Incorrect & Total \\
        \midrule
        OSS & Less Urgent (Older) & 119 & 68 & 187 \\
        OSS & More Urgent (Older) & 123 & 93 & 216 \\
        MedGemma-SFT & Less Urgent (Older) & 130 & 57 & 187 \\
        MedGemma-SFT & More Urgent (Older) & 152 & 64 & 216 \\
        \bottomrule
    \end{tabular}
    \caption{Prediction performance stratified by urgency level for older patients. MedGemma-SFT consistently outperforms the OSS model across both less urgent and more urgent cases, with higher correct predictions and fewer incorrect classifications.}
    \label{tab:age_order_older}
\end{table*}

We conduct a brief error analysis comparing two models, UrgentSFT-MedGemma and GPT-OSS-120B, on the publicly available PMR-Synth dataset. These models were selected to represent, respectively, a top-performing model in this task and a widely used open-source baseline. We focus our analysis on PMR-Synth to ensure reproducibility: the dataset includes structured EHR data with extractable demographic attributes relevant to our analysis and can be publicly released.

We begin our error analysis by examining the overlap in prediction outcomes between the two models, as shown in Table \ref{tab:prediction_overlap}. While a majority of instances are either correctly or incorrectly classified by both models, MedGemma-SFT accurately predicts substantially more samples that are inaccurately predicted by the OSS model than vice versa.

Next, we investigate whether prediction errors are systematically associated with demographic factors. Specifically, we test for correlations between gender and age. Table \ref{tab:urgency_gender} presents prediction performance stratified by urgency role and gender for both models. Across both OSS and MedGemma-SFT, accuracy is consistently higher when the more urgent case is male compared to female. In contrast, when the less urgent case is male, accuracy drops substantially—most notably for OSS, where correctness declines to 38.2\%. While MedGemma-SFT achieves higher accuracy across all urgency–gender combinations and exhibits a reduced gender disparity relative to OSS, lower performance persists in scenarios where the less urgent patient is male, indicating that gender-associated effects are not fully mitigated.

Next, we examine whether prediction correctness is associated with patient age ordering in the urgency comparison task. Table~\ref{tab:age_order_older} summarizes prediction performance when either the more urgent or less urgent patient is older (i.e., when patient A has a higher age than patient B in a given pair). For OSS, a chi-square test of independence found no statistically significant association between prediction correctness and age ordering ($p = 0.565$), with a very small effect size (Cramér’s $V = 0.03$). A similar pattern was observed for MedGemma-SFT, where correctness was also independent of age ordering ($p = 0.164$, Cramér’s $V = 0.07$). Overall, these results indicate that MedGemma-SFT’s prediction accuracy is largely unaffected by whether the older patient appears in the more urgent or less urgent role, suggesting minimal age-related bias in urgency-based decision-making.

\section{Prompts}

In Figures \ref{prompt:system_prompt}, \ref{prompt:urgent_reward} and \ref{prompt:urgent_sft} we show the core prompts used in this study. 

\begin{figure*}[!h]
    \centering
    \small
    \obeylines\obeyspaces
    \begin{promptbox}{System Prompt}
\#\#\# Role\: You are a medical expert. You specialize in understanding the urgency of medical queries.

\#\#\# Definitions: Triage nurses can categorize patients into 1 / 6 categories, where 1 is "most urgent" and 6 is "least urgent". 
- Level 1 --> Patient has life-threatening issue that needs immediate attention. 
- Level 2 --> Patient has non life-threatening issue that would benefit from same-day treatment (e.g. urgent care)
- Level 3 --> Patient should make an appointment with a doctor soon (1-3 days). 
- Level 4 --> Patient should see their doctor sometime in the near future (could be more than 3 days)
- Level 5 --> Patient has symptoms that can be treated at home, and would benefit from a message instructing them on what to do. 
- Level 6 --> Patient presents something that is a non-issue and no further steps are needed. \\

You can use this context to help think about which of two patients are more urgent. \\

\#\#\# Important Note: Patient messages may or may not be presented alongside structured EHR information such as medications, diagnoses, problem list, or demographics. These fields when available may contribute to medical urgency. \\

\#\#\# Relevant Context: Use the following information about medical urgency to help guide your response:
- Patients who are more medically urgent will benefit more from priority medical care. 
- Sometimes, a patient may be lower urgency because they have already been seen by a physician for this exact issue very recently. 
- Sometimes, the patient writes a message in a tone which is more urgent than their actual medical issue. Medical urgency must be based on objective medical content, not tone. E.g. health anxiety is not medically urgent. 
- Sometimes, a patient with a more severe chronic issue may be less urgent than a patient with a less severe acute issue, as the chronic patient may not have a time-sensitive issue. 
- Consider the risk factors of each patient. Two patients with the same symptoms may have different needs based on demographics or prior diagnosis that make them higher risk patients.
\end{promptbox}
    \caption{Shown is the system prompt provided to all models in all experiments}
    \label{prompt:system_prompt}
\end{figure*}

\begin{figure*}[!h]
    \centering
    \small
    \obeylines\obeyspaces
    \begin{promptbox}{UrgentSFT/Baseline Prompt}
\#\#\# CONTEXT \#\#\# 

You are a triage nurse who excels at determining the medical urgency of patient messages. 
There is a message already in your inbox from "Existing Patient". \\

A new message from "New Patient" has just arrived. \\

\#\#\# Existing Patient: \{ message 1 \}\\

\#\#\# New Patient: \{ message 2 \}\\

\#\#\# Instruction: Your job is to answer the question "Is the new message from "New Patient" *more* medically urgent than the existing message from "Existing Patient"? 
If the answer is YES, then "New Patient" will be seen by a doctor first. If the answer is NO, "Existing Patient" will be seen first. 

\#\#\# Response Format: Output "YES" or "NO" and nothing else. 

\end{promptbox}
    \caption{The prompt used for UrgentSFT, Instruct, and Reasoning Baselines. }
    \label{prompt:urgent_sft}
\end{figure*}

\begin{figure*}[!h]
    \centering
    \small
    \obeylines\obeyspaces
    \begin{promptbox}{UrgentReward}
\#\#\# Instruction: You are provided with a patient message sent to a clinician.
Your job is to generate a new patient message that is **more medically urgent** than the provided patient message.

Output the **more urgent** patient message and nothing else.

\#\#\# Patient Message: \{ message \}

\#\#\# More Urgent Patient Message:

\end{promptbox}
    \caption{The prompt used for UrgentReward. This prompt differs as we leverage pre-trained reward models which are trained to score completions. We thus re-formulate the task to better utilize existing knowledge.}
    \label{prompt:urgent_reward}
\end{figure*}

\end{document}